\documentclass{article}

    \PassOptionsToPackage{numbers, compress}{natbib}

    \usepackage[preprint]{neurips_2025}

\PassOptionsToPackage{dvipsnames,table}{xcolor}

\usepackage[utf8]{inputenc} 
\usepackage[T1]{fontenc}    
\usepackage{hyperref}       
\usepackage{url}            
\usepackage{booktabs}       
\usepackage{amsfonts}       
\usepackage{nicefrac}       
\usepackage{microtype}      
\usepackage{adjustbox}
\usepackage{graphicx}
\usepackage{tcolorbox}

\title{MedBookVQA: A Systematic and Comprehensive Medical Benchmark Derived from Open-Access Book}

\author{
Sau Lai Yip$^1$\footnotemark[1],~
Sunan He$^1$\footnotemark[1],~
Yuxiang Nie$^1$,~ 
Shu Pui Chan$^1$,~ \\
\textbf{Yilin Ye$^1$,}~
\textbf{Sum Ying Lam$^1$,}~
\textbf{Hao Chen$^{123}$}\footnotemark[2]\\
\small{$^1$Department of Computer Science and Engineering, The Hong Kong University of Science and Technology} \\
\small{$^2$Department of Chemical and Biological Engineering, The Hong Kong University of Science and Technology} \\
\small{$^3$Division of Life Science, The Hong Kong University of Science and Technology} \\
\texttt{slyipae}, \texttt{shebd}, \texttt{ynieae}, \texttt{spchanae}, \texttt{yyeaz}, \texttt{sylamau@connect.ust.hk} \\
\texttt{jhc@cse.ust.hk} \\
}

\newtcolorbox{promptenv}[1]{
        boxrule = 1.5pt,
        fontupper = \small,
        fonttitle = \bf\color{black},
        arc = 5pt,
        rounded corners,
        colframe = black,
        colbacktitle = white!97!gray,
        colback = white!97!gray,
        title = #1,
}

\begin{document}
\footnotetext[1]{Equal Contribution.}
\footnotetext[2]{Corresponding author.}
\maketitle
\begin{abstract}
The accelerating development of general medical artificial intelligence (GMAI), powered by multimodal large language models (MLLMs), offers transformative potential for addressing persistent healthcare challenges, including workforce deficits and escalating costs.
The parallel development of systematic evaluation benchmarks emerges as a critical imperative to enable performance assessment and provide technological guidance.
Meanwhile, as an invaluable knowledge source, the potential of medical textbooks for benchmark development remains underexploited.
Here, we present \textbf{MedBookVQA}, a systematic and comprehensive multimodal benchmark derived from open-access medical textbooks.
To curate this benchmark, we propose a standardized pipeline for automated extraction of medical figures while contextually aligning them with corresponding medical narratives.
Based on this curated data, we generate 5,000 clinically relevant questions spanning modality recognition, disease classification, anatomical identification, symptom diagnosis, and surgical procedures.
A multi-tier annotation system categorizes queries through hierarchical taxonomies encompassing medical imaging modalities (\textbf{42 categories}), body anatomies (\textbf{125 structures}), and clinical specialties (\textbf{31 departments}), enabling nuanced analysis across medical subdomains.
We evaluate a wide array of MLLMs, including proprietary, open-sourced, medical, and reasoning models, revealing significant performance disparities across task types and model categories. 
Our findings highlight critical capability gaps in current GMAI systems while establishing textbook-derived multimodal benchmarks as essential evaluation tools.
MedBookVQA establishes textbook-derived benchmarking as a critical paradigm for advancing clinical AI, exposing limitations in GMAI systems while providing anatomically structured performance metrics across specialties.
All data and code are available at \href{https://huggingface.co/datasets/slyipae1/MedBookVQA}{https://huggingface.co/datasets/slyipae1/MedBookVQA} and \href{https://github.com/slyipae1/MedBookVQA}{https://github.com/slyipae1/MedBookVQA}.
\end{abstract}

\begin{figure}[!th]
\centering
\includegraphics[width=1\textwidth]{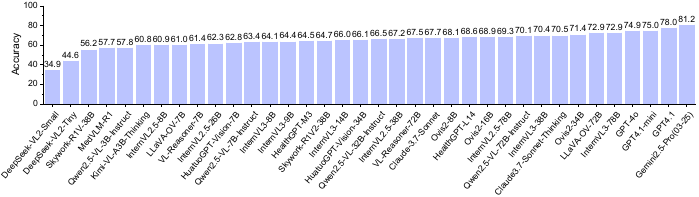}
\caption{Overall Performance of Multi-Modal Large Language Models on MedBookVQA.}
\label{AllOnly}
\end{figure}

\section{Introduction}
The global healthcare system confronts escalating challenges characterized by workforce deficiencies and unsustainable cost improvements, and artificial intelligence (AI) presents a transformative opportunity to address these systemic pressures~\cite{bajwa2021artificial}.
While earlier efforts focused on developing models for specific tasks or contexts \cite{bajwa2021artificial}, the emergence of general medical artificial intelligence (GMAI) systems marks a paradigm shift toward unified systems capable of addressing heterogeneous demands in medicine.
Multi-modal language models (MLLMs), with their capabilities in interpreting multimodal inputs across broad medical modalities and handling various tasks in diverse medical scenarios, serve as the cornerstone of GMAI systems. 
With the development of general-domain MLLMs, recent years have also seen the emergence of medically specialized MLLMs \cite{chen2024huatuogptvisioninjectingmedicalvisual, he2024gscogeneralizableaimedicine, lin2025healthgptmedicallargevisionlanguage, moor2023medflamingomultimodalmedicalfewshot}.
Alongside the advancement of these GMAI systems, it is equally important to construct comprehensive benchmarks to evaluate their performance, identify shortcomings, and provide technical guidance. 

Existing medical visual question answering (VQA) benchmarks and datasets have primarily targeted specific medical tasks~\cite{he2020pathvqa30000questionsmedical, liu2021slakesemanticallylabeledknowledgeenhanceddataset}, often revealing substantial limitations in both scope and applicability.
Recently, more diverse benchmarks encompassing multiple medical modalities have emerged, with some efforts aggregating and reformatting classification datasets~\cite{chen2024gmaimmbenchcomprehensivemultimodalevaluation, hu2024omnimedvqanewlargescalecomprehensive}.
However, these classification datasets, typically designed for single-use tasks, pose challenges for aggregation, thereby constraining their diversity and comprehensiveness.
Another notable source of data is PubMed Central, a digital library of biomedical journal literature \cite{chen2024huatuogptvisioninjectingmedicalvisual, zhang2024pmcvqavisualinstructiontuning}.
While extensive, these research articles often reflect cutting-edge findings that, in many cases, remain exploratory or lack widespread clinical validation.
In contrast, medical textbooks represent a more promising avenue for constructing datasets for GMAI, offering scalability, diversity, and a high degree of credibility.
Despite their potential, this rich and high-quality resource remains underutilized, with most efforts to date being narrowly focused on specific domains (e.g., pathology~\cite{he2020pathvqa30000questionsmedical}) or constrained to text-only formats~\cite{wu2023pmcllamabuildingopensourcelanguage}.

To bridge these gaps, we propose a standardized pipeline designed to extract medical figures from textbooks and pair them with corresponding textual descriptions. 
Additionally, we introduce a novel pipeline for generating VQA-format datasets tailored to five broadly applicable medical task categories, including modality recognition, disease classification, anatomical identification, symptom diagnosis, and surgical procedures.
To ensure scalability and generalizability, our pipeline employs broadly applicable matching rules and semantic descriptions of key medical concepts, including different image categories, question types, and constraints for generation. 
This design allows general-domain MLLMs to be effectively adapted for specialized medical tasks. 
To ensure the quality of the generated data, we incorporate rigorous filtering procedures and manual verification. 
Furthermore, we present a hierarchical annotation system that organizes dataset entries by modality, anatomical structure, and medical department, facilitating flexible data management and enabling fine-grained analyses.
Leveraging these pipelines, we introduce \textbf{MedBookVQA}, a meticulously curated benchmark dataset derived from publicly available medical textbooks.
MedBookVQA excels in both diversity and completeness, which spans an impressive 42 medical imaging modalities, 125 anatomical structures, and 31 medical departments.

To evaluate the current state of general medical artificial intelligence, we benchmark MedBookVQA with diverse MLLMs, including four proprietary general-purpose series, six open-source general-purpose series, two open-source medical series, and five reasoning-focused series.
Our evaluation yields two critical insights: (1) substantial variation in performance across different Med-VQA tasks, particularly for those requiring advanced medical knowledge and cross-modal reasoning; (2) proprietary general MLLMs outperform other general MLLMs, even medical counterparts, while reasoning models do not show impressive advantages over medical tasks.

In summary, our contributions are threefold:
\begin{itemize}
\item \textbf{Adaptive Pipelines for Medical VQA Dataset Construction}. We introduce scalable and robust pipelines designed to parse medical textbooks, generate diverse VQA tasks, and organize datasets within a hierarchical framework, ensuring both scalability and flexibility.
\item \textbf{Introduction of MedBookVQA, a Comprehensive Benchmark}. We present MedBookVQA, a structured and diverse benchmark dataset encompassing five VQA task categories across 42 imaging modalities, 125 anatomical structures, and 31 medical departments, establishing a valuable resource for advancing medical artificial intelligence.
\item \textbf{Extensive Evaluation of MLLMs}. We conduct a thorough evaluation of a wide array of proprietary and open-source MLLMs—including general-purpose, medical-specific, and reasoning-focused models—shedding light on current limitations and identifying key directions for future advancements in GMAI.
\end{itemize}

\begin{table*}
\centering
\caption{Comparison with other medical VQA benchmarks and datasets.}
\adjustbox{width=1\linewidth}{
\begin{tabular}{l|ccccccc}
\hline
\textbf{Medical VQA Benchmark} & \textbf{Task} & \textbf{Modality} & \textbf{Anatomy} & \textbf{Dept.}  & \textbf {Size} & \textbf{Medical Domain} & \textbf{Source} \\ 
\hline
Path-VQA \cite{he2020pathvqa30000questionsmedical} & 7 & — & — & — & 32k & Pathology & Pathology textbooks, PEIR Digital Library \\ 
SLAKE \cite{liu2021slakesemanticallylabeledknowledgeenhanceddataset} & 10 & 3 & 39 & — & 14k & Radiology & MSD \cite{simpson2019largeannotatedmedicalimage}, Chestx-ray8 \cite{Wang_2017}, CHAOS \cite{KAVUR2021101950} \\ 
OmniMedVQA \cite{hu2024omnimedvqanewlargescalecomprehensive} & 5 & 12 & 20 & — & 128k & Mixed & 73 Classification datasets \\ 
PMC-VQA \cite{zhang2024pmcvqavisualinstructiontuning} & — & > 20 & — & — & 227k & Mixed & PubMed Central \\ 
PubMedVision \cite{chen2024huatuogptvisioninjectingmedicalvisual} & — & > 9 & > 12 & — & 1.3m & Mixed & PubMed Central \\ 
GMAI-MMBench \cite{chen2024gmaimmbenchcomprehensivemultimodalevaluation} & 18 & 32 & — & 52 & 26K & Mixed & 284 datasets from public and hospitals \\ 
\hline 
MedBookVQA & 5 & 42 & 125 & 31 & 5k & Mixed & Medical books from DOAB \cite{doab} \\ 
\hline 
\vspace{-10pt}
\end{tabular} 
\label{tab:medical_benchmark} } 
\end{table*}

\begin{table*}
\centering
\caption{Comparison with other existing reasoning benchmarks.}
\adjustbox{width=1\linewidth}{
\begin{tabular}{l|cccccc}
\hline
\textbf{Reasoning Benchmark}  & \textbf {Size} & \textbf{Task} & \textbf{Domain} & \textbf{Source} \\ \hline
MMMU\cite{yue2024mmmumassivemultidisciplinemultimodal} & 11.5K & -- & Six college-level core disciplines & College exams, quizzes, and textbooks \\
MathVista\cite{lu2024mathvistaevaluatingmathematicalreasoning} & 15k & 7 & Math & 28 existing and 3 newly created multimodal datasets \\
MathVerse\cite{zhang2024mathversedoesmultimodalllm} & 2,612 & 12 & Math & Existing datasets and public question repositories \\
MATH-Vision\cite{wang2024measuring} & 3,040 & 16 & Math & Math competitions \\
MLLM-CompBench\cite{kil2025mllmcompbenchcomparativereasoningbenchmark} & 40k & 8 & General comparative reasoning & Public vision datasets\\
EMMA\cite{hao2025mllmsreasonmultimodalityemma} & 2,788 & 19 & Math, physics, chemistry, and coding & Public datasets, manual curation\\
MicroVQA\cite{burgess2025microvqamultimodalreasoningbenchmark} & 1,042 & 3 & Microscopy-based scientific research & Expert curation \\
\hline
MedBookVQA & 5k & 5 & Medical & Medical books from DOAB\cite{doab} \\ \hline
\vspace{-10pt}
\end{tabular}
\label{tab:reasoning_benchmark}
}
\end{table*}

\section{Related Work}

\textbf{Medical VQA Benchmarks and Datasets}.
Benchmarks of medical visual question answering have witnessed significant advancements through various dataset construction methodologies.
Path-VQA~\cite{he2020pathvqa30000questionsmedical} pioneered semi-automatic generation from pathology materials, employing linguistic transformations through verb decomposition and part-of-speech similarity analysis. 
While PMC-VQA~\cite{wu2023pmcllamabuildingopensourcelanguage} shares the textbook-based approach, it remains confined to textual knowledge extraction without visual integration. 
SLAKE~\cite{liu2021slakesemanticallylabeledknowledgeenhanceddataset} distinguishes itself through bilingual support and semantic knowledge enhancement in radiology, employing expert-designed question templates for anatomical specificity.
Recent large-scale efforts demonstrate divergent strategies: OmniMedVQA~\cite{hu2024omnimedvqanewlargescalecomprehensive} aggregates 73 classification datasets through templated multi-choice generation augmented by ChatGPT-3.5 distractor creation, whereas PMC-VQA~\cite{zhang2024pmcvqavisualinstructiontuning} leverages PubMed Central's biomedical corpus with generative model assistance. 
PubMedVision~\cite{chen2024huatuogptvisioninjectingmedicalvisual} addresses caption-based limitations through vision-aware refinement, expanding QA scenarios across eight clinical dimensions. 
The comprehensive GMAI-MMBench \cite{chen2024gmaimmbenchcomprehensivemultimodalevaluation} establishes new breadth with 284 source datasets, implementing a lexico-semantic hierarchy for granular evaluation across 38 modalities and 18 clinical specialties.
Our benchmark introduces several innovations compared to these medical datasets. Firstly, we employ public medical books to construct a general medical VQA, using an efficient pipeline applicable to any medical literature. Most general medical VQA datasets are constructed by reformatting existing datasets or leveraging PubMed Central. Although some datasets \cite{he2020pathvqa30000questionsmedical, wu2023pmcllamabuildingopensourcelanguage} also use books, they are often limited to small-scale, single-subject applications or rely solely on textual information. Secondly, our benchmark provides a broader coverage of modalities, anatomical structures, and departments owing to the diversity of the data source. Thirdly, its hierarchical labeling system facilitates flexible evaluation using specific subsets and fine-grained analysis. 

\textbf{Multi-modal Reasoning Benchmarks and Techniques}.
Existing multimodal benchmarks focus on specific domains or reasoning dimensions. 
MMMU \cite{yue2024mmmumassivemultidisciplinemultimodal} includes questions for six core disciplines, sourced from college education materials, to assess domain-specific perception and reasoning. 
MathVista \cite{lu2024mathvistaevaluatingmathematicalreasoning}, 
MathVerse \cite{zhang2024mathversedoesmultimodalllm}, and Math-Vision \cite{wang2024measuring} are mathematics benchmarks addressing various problem types. 
MLLM-CompBench \cite{kil2025mllmcompbenchcomparativereasoningbenchmark} offers 40k comparative reasoning VQA curated from public vision datasets across eight dimensions. 
EMMA \cite{hao2025mllmsreasonmultimodalityemma} presents 3k VQA requiring cross-modality reasoning across four subjects. 
MicroVQA \cite{burgess2025microvqamultimodalreasoningbenchmark} is an expert-curated benchmark for evaluating reasoning in microscopy-based scientific research.
To our knowledge, we are pioneering the introduction of a reasoning benchmark in the general medical domain.

\textbf{Multimodal Large Language Models}.
In recent years, MLLMs have emerged as powerful tools for general-domain multimodal tasks. Building on general LLMs and MLLMs, efforts have focused on developing medically specialized MLLMs for various tasks like question answering and report generation across medical domains. Notable examples include HealthGPT \cite{lin2025healthgptmedicallargevisionlanguage}, Huatuo-GPT Vision \cite{chen2024huatuogptvisioninjectingmedicalvisual}, MedDr \cite{he2024gscogeneralizableaimedicine}, and Med-Flamingo \cite{moor2023medflamingomultimodalmedicalfewshot}. 
There is also a growing interest in multimodal reasoning. Techniques such as Chain-of-Thought (CoT) reasoning have been extended to multimodal contexts as Multimodal Chain-of-Thought (MCOT) \cite{wang2025multimodalchainofthoughtreasoningcomprehensive}. 
Following the increasing popularity of reasoning LLMs \cite{deepseekai2025deepseekr1incentivizingreasoningcapability}, several reasoning MLLMs are being developed in both general \cite{peng2025skyworkr1vpioneeringmultimodal, chris2025skyworkr1v2multimodalhybrid, vl-rethinker, kimiteam2025kimivltechnicalreport} and medical \cite{pan2025medvlmr1incentivizingmedicalreasoning} domains.

\section{Benchmark Construction}
\subsection{Data Collection}
\label{set: Data Collection}
\begin{figure}
  \centering
  \includegraphics[width=1\textwidth]{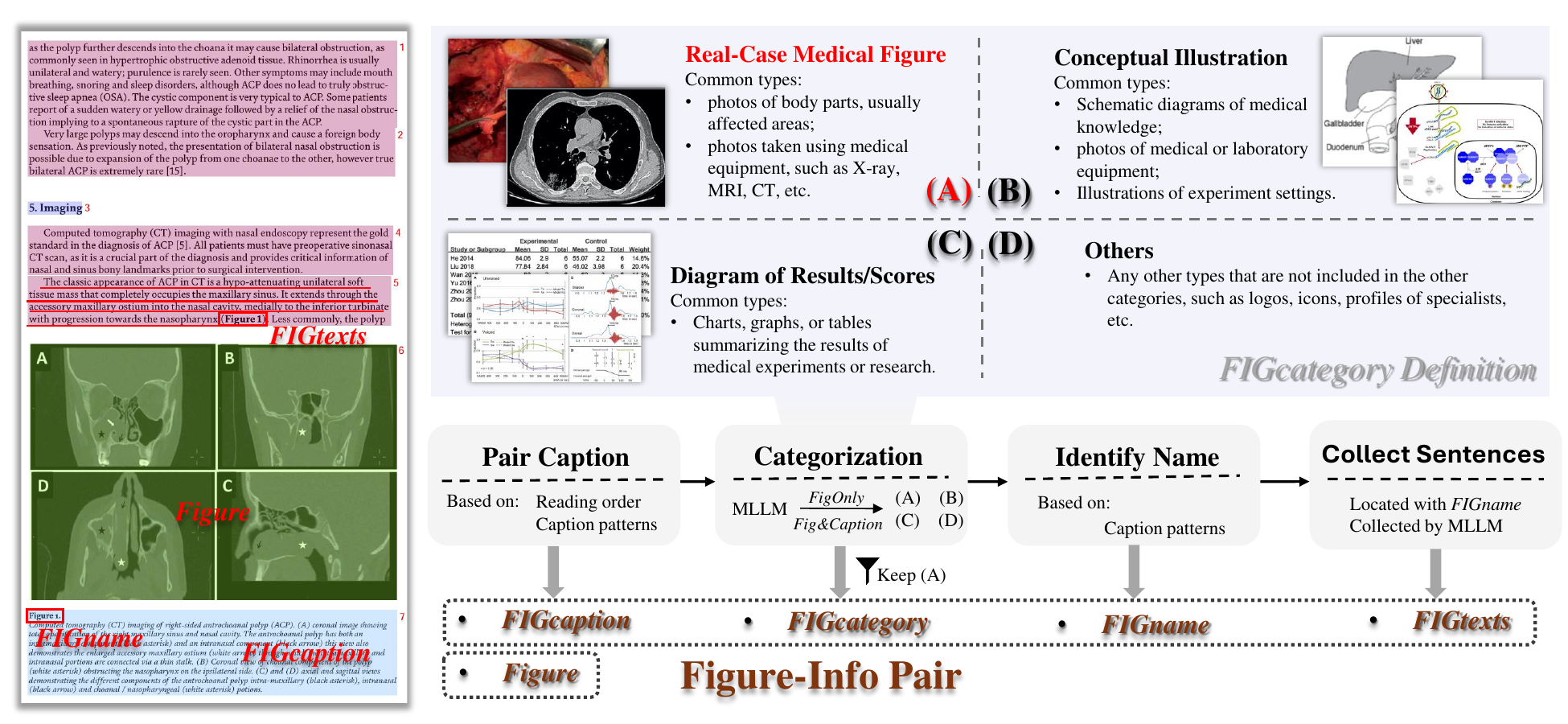}
  \caption{Pipeline for data collection. Medical figure-information pairs are extracted from the preprocessed layout result in four steps: (1) Captions are paired with images. (2) Images are categorized into four types. (3) Figure reference names are identified from the captions. (4) Sentences that describe the image by explicitly mentioning the figure reference name are collected.}
  \label{DataCollection}
\end{figure}

Our medical books are sourced from the Directory of Open Access Books (DOAB) \cite{doab} by querying its search engine using predefined medical keywords. MinerU \cite{wang2024mineruopensourcesolutionprecise} is used to process the downloaded books by dividing each page into smaller regions and labeling them by type (e.g., image, text, title). It also pairs images with their corresponding captions (if identified) and organizes regions according to reading order. Medical figure-information pairs are then obtained from the pre-processed results, each consisting of the figure (\textbf{\textit{Figure}}), figure captions (\textbf{\textit{FIGcaption}}), figure categories (\textbf{\textit{FIGcategory}}), figure reference names (\textbf{\textit{FIGname}}), and sentences in paragraphs that describe the image by explicitly mentioning the name (\textbf{\textit{FIGtexts}}). Figure \ref{DataCollection} illustrates this process.

\textbf{\textit{Figure} \& \textit{FIGcaption}} 
Although MinerU \cite{wang2024mineruopensourcesolutionprecise} attempts to pair captions with images, some failures occur because captions can be misclassified as text blocks. To address this, we designed matching rules based on patterns to recover captions, as captions typically begin with reference names like "Figure 1.1" or "Fig. 2". Unpaired images in the raw results are matched with captions from adjacent region blocks. Image and caption are uniquely paired.

\textbf{\textit{FIGcategory}} 
Successfully paired images are categorized using InternVL2-8B \cite{chen2024internvl} into four categories: 
\begin{itemize} 
    \item \textbf{Real-case medical figure}: Photos of body parts, usually affected areas, and images captured by medical equipment such as CT and MRI. 
    \item \textbf{Conceptual illustration}: Schematic diagrams of medical knowledge, photos of medical or laboratory equipment, illustrations of experiment settings, etc. 
    \item \textbf{Diagram of results or scores}: Charts, graphs, or tables summarizing the results of medical experiments or research. 
    \item \textbf{Other}: Any other type not listed above (e.g., logos, icons, etc.). 
\end{itemize} 
Categorization is performed using two approaches: (a) \textit{FigOnly}: inputting the figure only and (b) \textit{Fig\&Caption}: inputting both the figure and its caption. Images categorized as category A in either approach are retained.

\textbf{\textit{FIGname} \& \textit{FIGtexts}}
The \textit{FIGname} is extracted from the caption using pattern-matching rules similar to those used for caption recovery. This name is then used to locate sentences that describe the image by explicitly mentioning \textit{FIGname} in the surrounding paragraphs, namely \textit{FIGtexts}. InternVL2-8B \cite{chen2024internvl} is prompted to identify such sentences on the previous, current, and subsequent pages. Images with insufficient textual information (that is, the total words of \textit{FIGcaption} and \textit{FIGtexts} being less than 5) are excluded.

A total of 333,920 images were collected from 8,090 books, with 43,556 (around 13\%) classified as real-case medical images. After excluding erroneous entries (that is, those lacking \textit{FIGcaption}, \textit{FIGname}, or sufficient textual information), 36,820 valid figure-information pairs were obtained.

\subsection{VQA Generation and Filtering}
\label{set: VQA Generation and Filtering}
\begin{figure}
  \centering
  \includegraphics[width=1\textwidth]{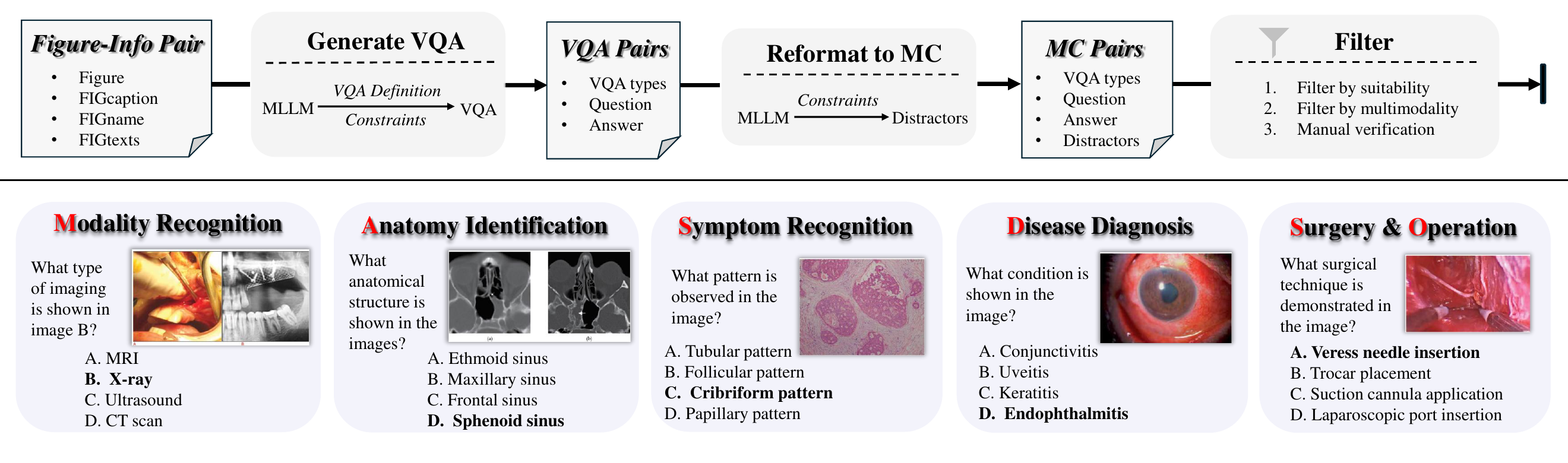}
  \caption{Pipeline for benchmark construction. The benchmark is constructed from figure-information pairs in three steps: (1) Five types of VQAs are generated from figure-information pairs. (2) The VQAs are reformatted into multiple-choice questions (MCs) by generating distractors. (3) Three filtering steps are performed according to suitability, multimodality, and manual verification.}
  \label{BenchmarkPipeline}
\end{figure}

The collected figure-information pairs are used to generate VQA, which are then filtered and labeled with a hierarchical system to form our MedBookVQA benchmark. Figure \ref{BenchmarkPipeline} presents the construction pipeline and sample VQAs from the benchmark.

\textbf{Generate VQA}
InternVL2.5-78B \cite{internvl2.5} is prompted to generate up to 6 VQAs for each figure-information pair from five predefined VQA types: 
\begin{itemize} 
    \item \textbf{(M) Modality Recognition}: Identify the image modality (e.g., MRI, CT, X-ray, etc.). 
    \item \textbf{(A) Anatomy Identification}: Identify the body structure or organ shown in the image. 
    \item \textbf{(S) Symptom Diagnosis}: Determine the symptom or abnormality (e.g., description, position) depicted in the image. 
    \item \textbf{(D) Disease Recognition}: Determine the disease or condition shown in the image. 
    \item \textbf{(SO) Surgery \& Operation}: Identify surgical actions, techniques, steps, workflows, use of tools or instruments, treatment, or positional arrangement of the patient in surgical images. 
\end{itemize} 
The prompt includes constraints to ensure concise answers in a few words or a phrase while discouraging VQAs unrelated to the image or requiring the textual information to answer. After generation, 10,000 VQAs are randomly selected for further processing. 

\textbf{Reformat VQA to MC}
The selected VQAs are reformatted into multiple-choice questions (MC) by generating distractors using Qwen-VL-Max \cite{qwen2.5}. Constraints in the prompt encourage challenging distractors of similar length and structure while discouraging the use of left-right distinctions, superordinate categories, and neighboring organs to avoid multiple correct answers.

\textbf{Filter}
Three steps are taken to filter the VQAs. First, Qwen-VL-Max \cite{qwen2.5} assesses suitability by excluding non-suitable types identified by manual inspection, including: (a) subjects not related to human disease diagnosis (e.g., veterinary medicine, forensic medicine, microbiology, plant pathology, etc.); (b) cases with inconsistency among the image, textual information, and VQA. The second step selects VQAs that require multimodality (i.e., the image) to answer. The question and choices are input to the LLM DeepSeek-R1 \cite{deepseekai2025deepseekr1incentivizingreasoningcapability} to determine the answerability without the image. Those deemed answerable are excluded as the images are redundant. Lastly, manual verification is performed to exclude: (1) cases with incorrect answers, (2) cases with answers appearing as textural annotations in the figures, and (3) choices with language shortcuts. The three filtering steps removed 2,065, 355, and 609 VQAs, respectively, resulting in 6,971 VQAs. For convenience, 1,000 were randomly selected for each VQA type to form the final benchmark, totaling 5,000 entries.

\subsection{Hierarchical Organization}
\begin{figure}
  \centering
  \includegraphics[width=0.9\textwidth]{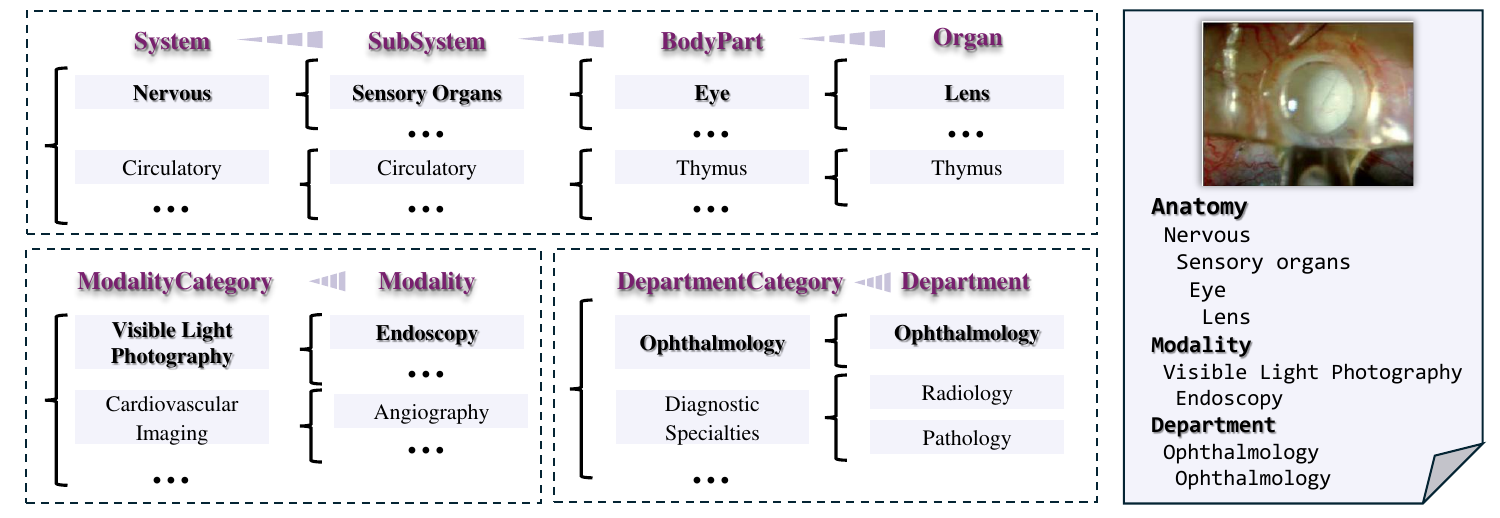}
  \caption{Hierarchical system of labels in terms of anatomy, modality, and department. A four-level tree is used for anatomy, while a two-level tree is used for modality and department, respectively.}
  \label{LabelSystem}
\end{figure}

Following the practice in \cite{chen2024gmaimmbenchcomprehensivemultimodalevaluation}, a hierarchical system (Figure \ref{LabelSystem}) labels each VQA in terms of anatomy, modality, and department. A four-level tree is constructed for anatomy and a two-level one for department and modality, respectively\footnote{Modality: article \cite{islam2023introductionmedicalimagingmodalities}, \url{https://en.wikipedia.org/wiki/Medical_imaging}} \footnote{Anatomy: \url{https://en.wikipedia.org/wiki/List_of_organs_of_the_human_body}} \footnote{Department: \url{https://www.mayoclinic.org/departments-centers}}. Qwen-VL-72B \cite{qwen2.5} labels each VQA at the lowest level. Particularly for anatomy, the second-to-last lowest level is used if no label can be assigned at the lowest level.

\section{Statistics and Distribution}

The MedBookVQA benchmark comprises 5,000 VQA with non-repeating images sourced from 1,103 distinct medical books. 
The most dominant modality is \textit{General Photo of Affected Area}, comprising 17.06\% of the data, followed closely by \textit{CT} and \textit{MRI} modalities, which together account for approximately 26.44\%. Other significant modalities include \textit{Visible Light Photography}, \textit{Radiography}, and \textit{Nuclear Imaging}, each contributing approximately 10\%.
The anatomical distribution reveals a focus on the \textit{Musculoskeletal} and \textit{Nervous} systems, collectively representing over 45\% of the data. Within the \textit{Musculoskeletal} category, body parts such as the \textit{Head}, \textit{Lower Limb}, and \textit{Spine} are prominent. The \textit{Nervous} system is predominantly represented by \textit{Sensory Organs}, particularly the \textit{Eye}, and \textit{Brain} structures. The \textit{Digestive} system follows, with a significant representation of \textit{Mouth} and \textit{Liver}, representing 17. 08\%. 
In terms of departmental categorization, diagnostic specialties such as \textit{Radiology} and \textit{Pathology} are prominent, comprising 16.4\%. \textit{Oral Health}, including \textit{Orthodontics} and \textit{Dentistry}, is also significantly represented with approximately 13\%. 
At the most granular level, the benchmark encompasses 42 modalities, 31 departments, and 125 organs. Overall, the benchmark is distinguished by its comprehensiveness, diversity, and hierarchically organized structure. More details can be found in the appendix.

\section{Experimental Evaluation}

The MedBookVQA benchmark is evaluate on a range of MLLMs to assess the current capabilities of GMAI. The evaluated models span the following four categories:
\begin{itemize}
    \item \textbf{Proprietary general MLLMs}: Gemini2.5-Pro (2025-03-26) \cite{GeminiPro2025}, GPT4.1 \cite{openai2024gpt4technicalreport}, GPT-4o \cite{openai2024gpt4technicalreport}, and Claude3.7-Sonnet \cite{Anthropic2024}.
    \item \textbf{Open-sourced general MLLMs}: InternVL3 \cite{zhu2025internvl3exploringadvancedtraining} and InternVL2.5 \cite{internvl2.5}, LLaVA-OV \cite{li2024llavaonevisioneasyvisualtask}, Ovis2 \cite{lu2024ovis}, Qwen2.5-VL \cite{qwen2.5}, and DeepSeek-VL2 \cite{wu2024deepseekvl2mixtureofexpertsvisionlanguagemodels}.
    \item \textbf{Open-sourced medical MLLMs}: HealthGPT \cite{lin2025healthgptmedicallargevisionlanguage}, and HuatuoGPT-Vision \cite{chen2024huatuogptvisioninjectingmedicalvisual}.
    \item \textbf{Reasoning MLLMs}: 
     \begin{itemize}
        \item Proprietary general reasoning MLLMs: Claude3.7-Sonnet-Thinking \cite{Anthropic2024}  
        \item Open-sourced general reasoning MLLMs: VL-Reasoner \cite{wang2025vlrethinkerincentivizingselfreflectionvisionlanguage}, Skywork-R1V \cite{peng2025skyworkr1vpioneeringmultimodal, chris2025skyworkr1v2multimodalhybrid}, and Kimi-VL-A3B-Thinking \cite{kimiteam2025kimivltechnicalreport}
        \item Open-sourced medical reasoning MLLMs: MedVLM-R1 \cite{pan2025medvlmr1incentivizingmedicalreasoning}.
    \end{itemize}
\end{itemize}
Each MLLM is evaluated on the MC tasks with temperature set to zero. Performance is measured as the percentage of correctly answered questions. The results are presented in Figure \ref{AllOnly}, with more details in the appendix. 

\begin{figure}
\vspace{-20pt}
\centering
\includegraphics[width=0.9\textwidth]{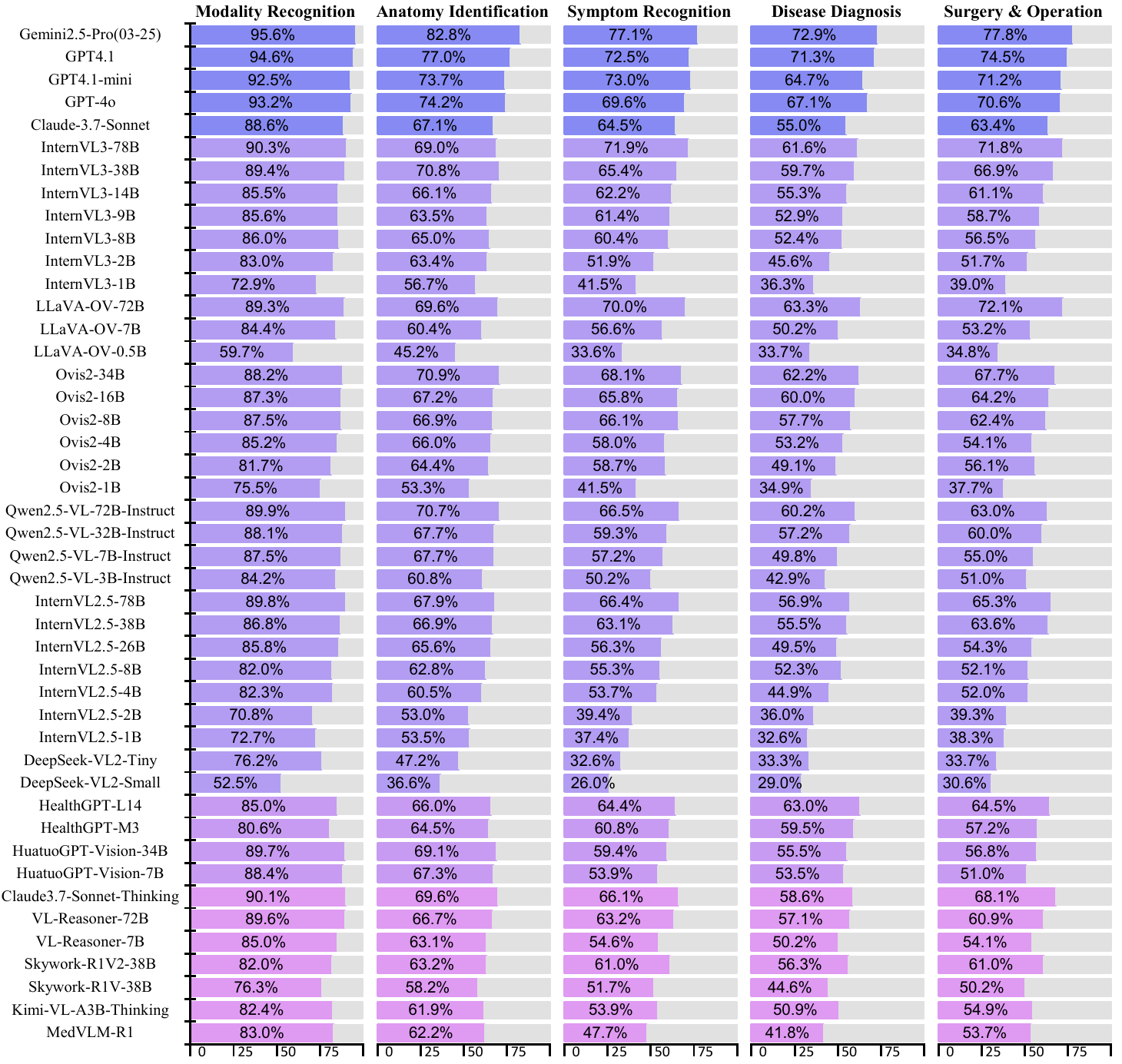}
\caption{Performance of MLLMs on the MedBookVQA benchmark, measured as the percentage of correctly answered questions across five VQA types, each comprising 1,000 questions (5,000 in total). Models are grouped into four categories. Within and across categories, models are ordered by descending performance, with models from the same series further sorted by size.}
\vspace{-20pt}
\label{AllResult}
\end{figure}
\subsection{Results and Analysis}
The evaluation result is analyzed in terms of VQA types and model categories. A more fine-grained analysis is further conducted on high-performing models and common medical aspects. 

\subsubsection{Performance across Types of VQA}

\textbf{VQAs that require strong medical knowledge and cross-modal clinical analysis are challenging to MLLMs across all categories} 
Although some state-of-the-art (SOTA) MLLMs in each category can surpass 79\% on all VQA tasks, high performance is predominantly observed in the \textit{Modality Recognition} type. In contrast, other types that demand stronger medical knowledge and cross-modal clinical analysis remain challenging for most MLLMs. This performance imbalance is more pronounced in less advanced models. For instance, while Gemini2.5-Pro(2025-03-26) \cite{GeminiPro2025} exhibits a gap of only 22.70\% points between its highest and lowest task performances, models like InternVL3-8B \cite{chen2024internvl} show a much larger discrepancy, with gaps extending to 33.60\% points. This indicates a significant variance in task-specific capabilities, highlighting the need for more robust clinical reasoning integration in future models. 

\subsubsection{Performance across MLLMs}

A comparison of the top-performing models within each category reveals the following overall performance hierarchy: \textbf{Proprietary general MLLMs > Open-sourced general MLLMs > Open-sourced medical MLLMs > Reasoning MLLMs} 

\textbf{(1) Proprietary general MLLMs demonstrate superior overall performance compared to other categories.} Among proprietary models, Gemini-2.5-pro \cite{GeminiPro2025} achieves the highest overall score of 81.24\%, excelling particularly in \textit{Modality Recognition} with a score of 95.60\%. This consistent performance across various VQA types underscores their versatility and adaptability in diverse scenarios, setting a benchmark for other models.

\textbf{(2) Open-sourced general MLLMs exhibit competitive performance, with some models approaching proprietary standards.} InternVL3-78B \cite{zhu2025internvl3exploringadvancedtraining} and Llava-OV-72B \cite{li2024llavaonevisioneasyvisualtask} both show strong performance, with overall scores of 72.92\% and 72.86\%, respectively. These models particularly excel in \textit{Anatomy Identification} and \textit{Surgery \& Operation}, indicating that open-sourced models are closing the gap with proprietary models in specialized medical areas. However, the performance gap of 8.32\% between the best open-source model and Gemini-2.5-pro \cite{GeminiPro2025} highlights the need for further advancements.

\textbf{(3) Open-sourced medical MLLMs show specialization in certain medical VQA types but lag in overall performance.} HealthGPT-L14 \cite{lin2025healthgptmedicallargevisionlanguage} and HuatuoGPT-Vision-34B \cite{chen2024huatuogptvisioninjectingmedicalvisual} are notable performers within this category, with scores of 68.58\% and 66.10\%, respectively. Their specialization in \textit{Disease Diagnosis} and \textit{Symptom Recognition }is evident, yet their overall performance is lower compared to top proprietary and open-sourced general models. This suggests that while they are tailored for specific medical applications, there is room for improvement in generalizability and adaptability across other VQA types.

\textbf{(4) Reasoning-enhanced MLLMs show inconsistent performance gains on medical tasks.} While Claude3.7-Sonnet-Thinking \cite{Anthropic2024} improves over its base version by 2.78\%, open-source reasoning models like VL-Reasoner \cite{vl-rethinker} and Skywork-R1V2-38B \cite{chris2025skyworkr1v2multimodalhybrid} underperform compared to general models. This indicates that current reasoning enhancements are insufficient for complex medical domains, necessitating further development in reasoning strategies and contextual understanding specific to medical tasks.

\subsubsection{Analysis by Anatomy, Modality, and Department}
\begin{figure}
\centering
\includegraphics[width=1\textwidth]{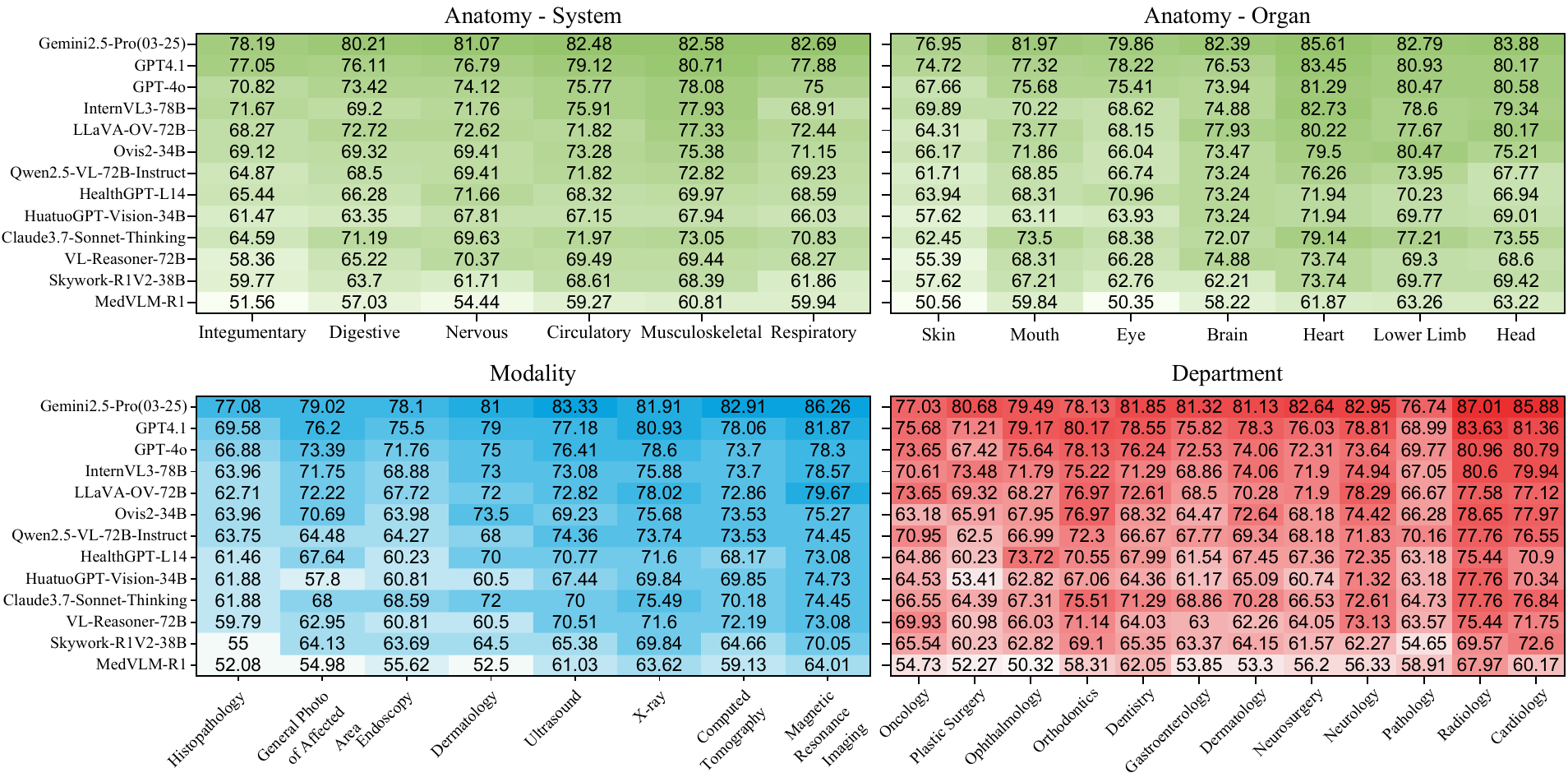}
\caption{Performance of high-performing MLLMs on common categories in anatomy, modality, and department.}
\label{DetailResult}
\vspace{-10pt}
\end{figure}

We select the highest-performing models from each category and common types (over 200 entries) of anatomy, modality, and department using the hierarchical label system to perform a more fine-grained analysis (see Figure \ref{DetailResult}). The ranking is generally consistent with the above conclusions. 
At the \textbf{anatomical System} level, most models show noticeable weaknesses in the \textit{Integumentary}, \textit{Respiratory}, and \textit{Digestive} systems. At the \textbf{Organ} level, there is evident weakness in \textit{Skin} and \textit{Eye} type problems for all the models. 
For \textit{Modality}, all models find problems from \textit{Histopathology} and \textit{Endoscopy} difficult. Those less-performing models find questioning from \textit{Dermatology} and \textit{Computed Tomography} more challenging. 
As for the \textit{Department}, an evident higher performance can be observed from most models for \textit{Radiology}, \textit{Cardiology}, and \textit{Neurology}, while the opposite case is shown for \textit{Pathology}, \textit{Plastic Surgery}, \textit{Gastroenterology}, and \textit{Neurosurgery}, especially for lower-performing models. 
Particularly, an evidently inconsistent low performance on \textit{General Photo of Affected Area} and \textit{Plastic Surgery} is observed for HuatuoGPT-Vision-34B \cite{chen2024huatuogptvisioninjectingmedicalvisual}. We hypothesize that this is due to the lack of such images in their training dataset PubMedVision \cite{chen2024huatuogptvisioninjectingmedicalvisual}, which is sourced from PubMed Central.

\section{Limitations and Future Work}

Despite our contributions, there are limitations to consider. 
As with other datasets, errors are inevitable in our data. 
The construction involved LLMs and MLLMs, which are susceptible to issues such as hallucination, as disclosed by other works. 
Although extensive experiments and analyses have been conducted on a small scale to minimize error rates during prompt design, the data has not been thoroughly verified by specialized experts due to the broad range of expertise required. 
Additionally, the use of such models for generative tasks, including VQA and distractor generation, may introduce linguistic shortcuts, potentially affecting the accurate assessment of the models' capabilities. 
Furthermore, our pipeline may not fully utilize the information in books, as figure-related information is not confined to direct neighboring references. The rich knowledge and local arrangements in such literature present promising opportunities for further exploration. 

\section{Conclusion}

The development of the MedBookVQA benchmark represents a significant advancement for GMAI by providing a comprehensive and fine-grained evaluation. Structured into five VQA types relevant to various medical domains and encompassing 42 modalities, 125 anatomical structures, and 31 departments, the benchmark enables a thorough assessment of model capabilities and generalizability in medical tasks. Its hierarchical organization further supports flexible testing and detailed analysis of specific tasks. Its potential to accelerate the development of robust GMAI systems is highly promising. In addition to the introduction of the benchmark, the effective pipelines developed for its construction address the current underutilization of medical books, which are invaluable sources of reliable medical knowledge.

\bibliographystyle{abbrvnat}
\medskip
{
\small
\bibliography{bibliography}

\begin{thebibliography}{42}
\providecommand{\natexlab}[1]{#1}
\providecommand{\url}[1]{\texttt{#1}}
\expandafter\ifx\csname urlstyle\endcsname\relax
  \providecommand{\doi}[1]{doi: #1}\else
  \providecommand{\doi}{doi: \begingroup \urlstyle{rm}\Url}\fi

\bibitem[doa()]{doab}
Directory of open access books.
\newblock URL \url{https://www.doabooks.org/}.
\newblock Accessed: 2024-11-05.

\bibitem[Anthropic(2024)]{Anthropic2024}
Anthropic.
\newblock Claude 3.7 sonnet system card.
\newblock \url{https://assets.anthropic.com/m/785e231869ea8b3b/original/claude-3-7-sonnet-system-card.pdf}, 2024.
\newblock Accessed: 2025-05-11.

\bibitem[Bajwa et~al.(2021)Bajwa, Munir, Nori, and Williams]{bajwa2021artificial}
J.~Bajwa, U.~Munir, A.~Nori, and B.~Williams.
\newblock Artificial intelligence in healthcare: Transforming the practice of medicine.
\newblock \emph{Future Healthcare Journal}, 8\penalty0 (2):\penalty0 e188--e194, 2021.
\newblock \doi{10.7861/fhj.2021-0095}.
\newblock URL \url{https://pmc.ncbi.nlm.nih.gov/articles/PMC8285156/}.

\bibitem[Burgess et~al.(2025)Burgess, Nirschl, Bravo-Sánchez, Lozano, Gupte, Galaz-Montoya, Zhang, Su, Bhowmik, Coman, Hasan, Johannesson, Leineweber, Nair, Yarlagadda, Zuraski, Chiu, Cohen, Hansen, Leonetti, Liu, Lundberg, and Yeung-Levy]{burgess2025microvqamultimodalreasoningbenchmark}
J.~Burgess, J.~J. Nirschl, L.~Bravo-Sánchez, A.~Lozano, S.~R. Gupte, J.~G. Galaz-Montoya, Y.~Zhang, Y.~Su, D.~Bhowmik, Z.~Coman, S.~M. Hasan, A.~Johannesson, W.~D. Leineweber, M.~G. Nair, R.~Yarlagadda, C.~Zuraski, W.~Chiu, S.~Cohen, J.~N. Hansen, M.~D. Leonetti, C.~Liu, E.~Lundberg, and S.~Yeung-Levy.
\newblock Microvqa: A multimodal reasoning benchmark for microscopy-based scientific research, 2025.
\newblock URL \url{https://arxiv.org/abs/2503.13399}.

\bibitem[Chen et~al.(2024{\natexlab{a}})Chen, Ouyang, Gao, Chen, Chen, Wang, Zhang, Cai, Ji, Yu, Wan, and Wang]{chen2024huatuogptvisioninjectingmedicalvisual}
J.~Chen, R.~Ouyang, A.~Gao, S.~Chen, G.~H. Chen, X.~Wang, R.~Zhang, Z.~Cai, K.~Ji, G.~Yu, X.~Wan, and B.~Wang.
\newblock Huatuogpt-vision, towards injecting medical visual knowledge into multimodal llms at scale, 2024{\natexlab{a}}.
\newblock URL \url{https://arxiv.org/abs/2406.19280}.

\bibitem[Chen et~al.(2024{\natexlab{b}})Chen, Ye, Wang, Li, Deng, Li, Li, Duan, Huang, Su, Wang, Zhang, Fu, Cai, Zhuang, Seibel, He, and Qiao]{chen2024gmaimmbenchcomprehensivemultimodalevaluation}
P.~Chen, J.~Ye, G.~Wang, Y.~Li, Z.~Deng, W.~Li, T.~Li, H.~Duan, Z.~Huang, Y.~Su, B.~Wang, S.~Zhang, B.~Fu, J.~Cai, B.~Zhuang, E.~J. Seibel, J.~He, and Y.~Qiao.
\newblock Gmai-mmbench: A comprehensive multimodal evaluation benchmark towards general medical ai, 2024{\natexlab{b}}.
\newblock URL \url{https://arxiv.org/abs/2408.03361}.

\bibitem[Chen et~al.(2024{\natexlab{c}})Chen, Wu, Wang, Su, Chen, Xing, Zhong, Zhang, Zhu, Lu, et~al.]{chen2024internvl}
Z.~Chen, J.~Wu, W.~Wang, W.~Su, G.~Chen, S.~Xing, M.~Zhong, Q.~Zhang, X.~Zhu, L.~Lu, et~al.
\newblock Internvl: Scaling up vision foundation models and aligning for generic visual-linguistic tasks.
\newblock In \emph{Proceedings of the IEEE/CVF Conference on Computer Vision and Pattern Recognition}, pages 24185--24198, 2024{\natexlab{c}}.

\bibitem[Chen et~al.(2025)Chen, Wang, Cao, Liu, Gao, Cui, Zhu, Ye, Tian, Liu, Gu, Wang, Li, Ren, Chen, Luo, Wang, Jiang, Wang, He, Shi, Zhang, Lv, Wang, Shao, Chu, Tu, He, Wu, Deng, Ge, Chen, Zhang, Wang, Dou, Lu, Zhu, Lu, Lin, Qiao, Dai, and Wang]{internvl2.5}
Z.~Chen, W.~Wang, Y.~Cao, Y.~Liu, Z.~Gao, E.~Cui, J.~Zhu, S.~Ye, H.~Tian, Z.~Liu, L.~Gu, X.~Wang, Q.~Li, Y.~Ren, Z.~Chen, J.~Luo, J.~Wang, T.~Jiang, B.~Wang, C.~He, B.~Shi, X.~Zhang, H.~Lv, Y.~Wang, W.~Shao, P.~Chu, Z.~Tu, T.~He, Z.~Wu, H.~Deng, J.~Ge, K.~Chen, K.~Zhang, L.~Wang, M.~Dou, L.~Lu, X.~Zhu, T.~Lu, D.~Lin, Y.~Qiao, J.~Dai, and W.~Wang.
\newblock Expanding performance boundaries of open-source multimodal models with model, data, and test-time scaling, 2025.
\newblock URL \url{https://arxiv.org/abs/2412.05271}.

\bibitem[Chris et~al.(2025)Chris, Wei, Peng, Wang, Qiu, Shen, Xie, Pei, Zhang, Hao, Song, Liu, and Zhou]{chris2025skyworkr1v2multimodalhybrid}
Chris, Y.~Wei, Y.~Peng, X.~Wang, W.~Qiu, W.~Shen, T.~Xie, J.~Pei, J.~Zhang, Y.~Hao, X.~Song, Y.~Liu, and Y.~Zhou.
\newblock Skywork r1v2: Multimodal hybrid reinforcement learning for reasoning, 2025.
\newblock URL \url{https://arxiv.org/abs/2504.16656}.

\bibitem[DeepSeek-AI(2025)]{deepseekai2025deepseekr1incentivizingreasoningcapability}
DeepSeek-AI.
\newblock Deepseek-r1: Incentivizing reasoning capability in llms via reinforcement learning, 2025.
\newblock URL \url{https://arxiv.org/abs/2501.12948}.

\bibitem[{Google DeepMind}(2025)]{GeminiPro2025}
{Google DeepMind}.
\newblock Gemini pro - google deepmind, 2025.
\newblock URL \url{https://deepmind.google/technologies/gemini/pro/}.
\newblock Accessed: 2025-05-11.

\bibitem[Hao et~al.(2025)Hao, Gu, Wang, Li, Yang, Wang, and Cheng]{hao2025mllmsreasonmultimodalityemma}
Y.~Hao, J.~Gu, H.~W. Wang, L.~Li, Z.~Yang, L.~Wang, and Y.~Cheng.
\newblock Can mllms reason in multimodality? emma: An enhanced multimodal reasoning benchmark, 2025.
\newblock URL \url{https://arxiv.org/abs/2501.05444}.

\bibitem[He et~al.(2024)He, Nie, Wang, Yang, Wang, Cai, Chen, Xu, Luo, Xiang, Lin, Wu, Peng, Shih, Xu, Wu, Wang, Chan, Vardhanabhuti, Chu, Zheng, Rajpurkar, Zhang, and Chen]{he2024gscogeneralizableaimedicine}
S.~He, Y.~Nie, H.~Wang, S.~Yang, Y.~Wang, Z.~Cai, Z.~Chen, Y.~Xu, L.~Luo, H.~Xiang, X.~Lin, M.~Wu, Y.~Peng, G.~Shih, Z.~Xu, X.~Wu, Q.~Wang, R.~C.~K. Chan, V.~Vardhanabhuti, W.~C.~W. Chu, Y.~Zheng, P.~Rajpurkar, K.~Zhang, and H.~Chen.
\newblock Gsco: Towards generalizable ai in medicine via generalist-specialist collaboration, 2024.
\newblock URL \url{https://arxiv.org/abs/2404.15127}.

\bibitem[He et~al.(2020)He, Zhang, Mou, Xing, and Xie]{he2020pathvqa30000questionsmedical}
X.~He, Y.~Zhang, L.~Mou, E.~Xing, and P.~Xie.
\newblock Pathvqa: 30000+ questions for medical visual question answering, 2020.
\newblock URL \url{https://arxiv.org/abs/2003.10286}.

\bibitem[Hu et~al.(2024)Hu, Li, Lu, Shao, He, Qiao, and Luo]{hu2024omnimedvqanewlargescalecomprehensive}
Y.~Hu, T.~Li, Q.~Lu, W.~Shao, J.~He, Y.~Qiao, and P.~Luo.
\newblock Omnimedvqa: A new large-scale comprehensive evaluation benchmark for medical lvlm, 2024.
\newblock URL \url{https://arxiv.org/abs/2402.09181}.

\bibitem[Islam et~al.(2023)Islam, Nasim, Hossain, Ullah, Gupta, and Bhuiyan]{islam2023introductionmedicalimagingmodalities}
S.~K. M.~S. Islam, M.~A.~A. Nasim, I.~Hossain, M.~A. Ullah, K.~D. Gupta, and M.~M.~H. Bhuiyan.
\newblock Introduction of medical imaging modalities, 2023.
\newblock URL \url{https://arxiv.org/abs/2306.01022}.

\bibitem[Kavur et~al.(2021)Kavur, Gezer, Barış, Aslan, Conze, Groza, Pham, Chatterjee, Ernst, Özkan, Baydar, Lachinov, Han, Pauli, Isensee, Perkonigg, Sathish, Rajan, Sheet, Dovletov, Speck, Nürnberger, Maier-Hein, {Bozdağı Akar}, Ünal, Dicle, and Selver]{KAVUR2021101950}
A.~E. Kavur, N.~S. Gezer, M.~Barış, S.~Aslan, P.-H. Conze, V.~Groza, D.~D. Pham, S.~Chatterjee, P.~Ernst, S.~Özkan, B.~Baydar, D.~Lachinov, S.~Han, J.~Pauli, F.~Isensee, M.~Perkonigg, R.~Sathish, R.~Rajan, D.~Sheet, G.~Dovletov, O.~Speck, A.~Nürnberger, K.~H. Maier-Hein, G.~{Bozdağı Akar}, G.~Ünal, O.~Dicle, and M.~A. Selver.
\newblock {CHAOS Challenge - combined (CT-MR) healthy abdominal organ segmentation}.
\newblock \emph{Medical Image Analysis}, 69:\penalty0 101950, Apr. 2021.
\newblock ISSN 1361-8415.
\newblock \doi{https://doi.org/10.1016/j.media.2020.101950}.
\newblock URL \url{http://www.sciencedirect.com/science/article/pii/S1361841520303145}.

\bibitem[Kil et~al.(2025)Kil, Mai, Lee, Wang, Cheng, Wang, Liu, Chowdhury, and Chao]{kil2025mllmcompbenchcomparativereasoningbenchmark}
J.~Kil, Z.~Mai, J.~Lee, Z.~Wang, K.~Cheng, L.~Wang, Y.~Liu, A.~Chowdhury, and W.-L. Chao.
\newblock Mllm-compbench: A comparative reasoning benchmark for multimodal llms, 2025.
\newblock URL \url{https://arxiv.org/abs/2407.16837}.

\bibitem[Li et~al.(2024)Li, Zhang, Guo, Zhang, Li, Zhang, Zhang, Zhang, Li, Liu, and Li]{li2024llavaonevisioneasyvisualtask}
B.~Li, Y.~Zhang, D.~Guo, R.~Zhang, F.~Li, H.~Zhang, K.~Zhang, P.~Zhang, Y.~Li, Z.~Liu, and C.~Li.
\newblock Llava-onevision: Easy visual task transfer, 2024.
\newblock URL \url{https://arxiv.org/abs/2408.03326}.

\bibitem[Lin et~al.(2025)Lin, Zhang, Li, Yuan, Yu, Li, He, Jiang, Li, Song, Tang, Xiao, Lin, Zhuang, and Ooi]{lin2025healthgptmedicallargevisionlanguage}
T.~Lin, W.~Zhang, S.~Li, Y.~Yuan, B.~Yu, H.~Li, W.~He, H.~Jiang, M.~Li, X.~Song, S.~Tang, J.~Xiao, H.~Lin, Y.~Zhuang, and B.~C. Ooi.
\newblock Healthgpt: A medical large vision-language model for unifying comprehension and generation via heterogeneous knowledge adaptation, 2025.
\newblock URL \url{https://arxiv.org/abs/2502.09838}.

\bibitem[Liu et~al.(2021)Liu, Zhan, Xu, Ma, Yang, and Wu]{liu2021slakesemanticallylabeledknowledgeenhanceddataset}
B.~Liu, L.-M. Zhan, L.~Xu, L.~Ma, Y.~Yang, and X.-M. Wu.
\newblock Slake: A semantically-labeled knowledge-enhanced dataset for medical visual question answering, 2021.
\newblock URL \url{https://arxiv.org/abs/2102.09542}.

\bibitem[Lu et~al.(2024{\natexlab{a}})Lu, Bansal, Xia, Liu, Li, Hajishirzi, Cheng, Chang, Galley, and Gao]{lu2024mathvistaevaluatingmathematicalreasoning}
P.~Lu, H.~Bansal, T.~Xia, J.~Liu, C.~Li, H.~Hajishirzi, H.~Cheng, K.-W. Chang, M.~Galley, and J.~Gao.
\newblock Mathvista: Evaluating mathematical reasoning of foundation models in visual contexts, 2024{\natexlab{a}}.
\newblock URL \url{https://arxiv.org/abs/2310.02255}.

\bibitem[Lu et~al.(2024{\natexlab{b}})Lu, Li, Chen, Xu, Luo, Zhang, and Ye]{lu2024ovis}
S.~Lu, Y.~Li, Q.-G. Chen, Z.~Xu, W.~Luo, K.~Zhang, and H.-J. Ye.
\newblock Ovis: Structural embedding alignment for multimodal large language model.
\newblock \emph{arXiv:2405.20797}, 2024{\natexlab{b}}.

\bibitem[Moor et~al.(2023)Moor, Huang, Wu, Yasunaga, Zakka, Dalmia, Reis, Rajpurkar, and Leskovec]{moor2023medflamingomultimodalmedicalfewshot}
M.~Moor, Q.~Huang, S.~Wu, M.~Yasunaga, C.~Zakka, Y.~Dalmia, E.~P. Reis, P.~Rajpurkar, and J.~Leskovec.
\newblock Med-flamingo: a multimodal medical few-shot learner, 2023.
\newblock URL \url{https://arxiv.org/abs/2307.15189}.

\bibitem[OpenAI et~al.(2024)OpenAI, Achiam, Adler, Agarwal, Ahmad, Akkaya, Aleman, Almeida, Altenschmidt, Altman, Anadkat, Avila, Babuschkin, Balaji, Balcom, Baltescu, Bao, Bavarian, Belgum, Bello, Berdine, Bernadett-Shapiro, Berner, Bogdonoff, Boiko, Boyd, Brakman, Brockman, Brooks, Brundage, Button, Cai, Campbell, Cann, Carey, Carlson, Carmichael, Chan, Chang, Chantzis, Chen, Chen, Chen, Chen, Chen, Chess, Cho, Chu, Chung, Cummings, Currier, Dai, Decareaux, Degry, Deutsch, Deville, Dhar, Dohan, Dowling, Dunning, Ecoffet, Eleti, Eloundou, Farhi, Fedus, Felix, Fishman, Forte, Fulford, Gao, Georges, Gibson, Goel, Gogineni, Goh, Gontijo-Lopes, Gordon, Grafstein, Gray, Greene, Gross, Gu, Guo, Hallacy, Han, Harris, He, Heaton, Heidecke, Hesse, Hickey, Hickey, Hoeschele, Houghton, Hsu, Hu, Hu, Huizinga, Jain, Jain, Jang, Jiang, Jiang, Jin, Jin, Jomoto, Jonn, Jun, Kaftan, Łukasz Kaiser, Kamali, Kanitscheider, Keskar, Khan, Kilpatrick, Kim, Kim, Kim, Kirchner, Kiros, Knight, Kokotajlo, Łukasz Kondraciuk, Kondrich,
  Konstantinidis, Kosic, Krueger, Kuo, Lampe, Lan, Lee, Leike, Leung, Levy, Li, Lim, Lin, Lin, Litwin, Lopez, Lowe, Lue, Makanju, Malfacini, Manning, Markov, Markovski, Martin, Mayer, Mayne, McGrew, McKinney, McLeavey, McMillan, McNeil, Medina, Mehta, Menick, Metz, Mishchenko, Mishkin, Monaco, Morikawa, Mossing, Mu, Murati, Murk, Mély, Nair, Nakano, Nayak, Neelakantan, Ngo, Noh, Ouyang, O'Keefe, Pachocki, Paino, Palermo, Pantuliano, Parascandolo, Parish, Parparita, Passos, Pavlov, Peng, Perelman, de~Avila Belbute~Peres, Petrov, de~Oliveira~Pinto, Michael, Pokorny, Pokrass, Pong, Powell, Power, Power, Proehl, Puri, Radford, Rae, Ramesh, Raymond, Real, Rimbach, Ross, Rotsted, Roussez, Ryder, Saltarelli, Sanders, Santurkar, Sastry, Schmidt, Schnurr, Schulman, Selsam, Sheppard, Sherbakov, Shieh, Shoker, Shyam, Sidor, Sigler, Simens, Sitkin, Slama, Sohl, Sokolowsky, Song, Staudacher, Such, Summers, Sutskever, Tang, Tezak, Thompson, Tillet, Tootoonchian, Tseng, Tuggle, Turley, Tworek, Uribe, Vallone, Vijayvergiya,
  Voss, Wainwright, Wang, Wang, Wang, Ward, Wei, Weinmann, Welihinda, Welinder, Weng, Weng, Wiethoff, Willner, Winter, Wolrich, Wong, Workman, Wu, Wu, Wu, Xiao, Xu, Yoo, Yu, Yuan, Zaremba, Zellers, Zhang, Zhang, Zhao, Zheng, Zhuang, Zhuk, and Zoph]{openai2024gpt4technicalreport}
OpenAI, J.~Achiam, S.~Adler, S.~Agarwal, L.~Ahmad, I.~Akkaya, F.~L. Aleman, D.~Almeida, J.~Altenschmidt, S.~Altman, S.~Anadkat, R.~Avila, I.~Babuschkin, S.~Balaji, V.~Balcom, P.~Baltescu, H.~Bao, M.~Bavarian, J.~Belgum, I.~Bello, J.~Berdine, G.~Bernadett-Shapiro, C.~Berner, L.~Bogdonoff, O.~Boiko, M.~Boyd, A.-L. Brakman, G.~Brockman, T.~Brooks, M.~Brundage, K.~Button, T.~Cai, R.~Campbell, A.~Cann, B.~Carey, C.~Carlson, R.~Carmichael, B.~Chan, C.~Chang, F.~Chantzis, D.~Chen, S.~Chen, R.~Chen, J.~Chen, M.~Chen, B.~Chess, C.~Cho, C.~Chu, H.~W. Chung, D.~Cummings, J.~Currier, Y.~Dai, C.~Decareaux, T.~Degry, N.~Deutsch, D.~Deville, A.~Dhar, D.~Dohan, S.~Dowling, S.~Dunning, A.~Ecoffet, A.~Eleti, T.~Eloundou, D.~Farhi, L.~Fedus, N.~Felix, S.~P. Fishman, J.~Forte, I.~Fulford, L.~Gao, E.~Georges, C.~Gibson, V.~Goel, T.~Gogineni, G.~Goh, R.~Gontijo-Lopes, J.~Gordon, M.~Grafstein, S.~Gray, R.~Greene, J.~Gross, S.~S. Gu, Y.~Guo, C.~Hallacy, J.~Han, J.~Harris, Y.~He, M.~Heaton, J.~Heidecke, C.~Hesse, A.~Hickey,
  W.~Hickey, P.~Hoeschele, B.~Houghton, K.~Hsu, S.~Hu, X.~Hu, J.~Huizinga, S.~Jain, S.~Jain, J.~Jang, A.~Jiang, R.~Jiang, H.~Jin, D.~Jin, S.~Jomoto, B.~Jonn, H.~Jun, T.~Kaftan, Łukasz Kaiser, A.~Kamali, I.~Kanitscheider, N.~S. Keskar, T.~Khan, L.~Kilpatrick, J.~W. Kim, C.~Kim, Y.~Kim, J.~H. Kirchner, J.~Kiros, M.~Knight, D.~Kokotajlo, Łukasz Kondraciuk, A.~Kondrich, A.~Konstantinidis, K.~Kosic, G.~Krueger, V.~Kuo, M.~Lampe, I.~Lan, T.~Lee, J.~Leike, J.~Leung, D.~Levy, C.~M. Li, R.~Lim, M.~Lin, S.~Lin, M.~Litwin, T.~Lopez, R.~Lowe, P.~Lue, A.~Makanju, K.~Malfacini, S.~Manning, T.~Markov, Y.~Markovski, B.~Martin, K.~Mayer, A.~Mayne, B.~McGrew, S.~M. McKinney, C.~McLeavey, P.~McMillan, J.~McNeil, D.~Medina, A.~Mehta, J.~Menick, L.~Metz, A.~Mishchenko, P.~Mishkin, V.~Monaco, E.~Morikawa, D.~Mossing, T.~Mu, M.~Murati, O.~Murk, D.~Mély, A.~Nair, R.~Nakano, R.~Nayak, A.~Neelakantan, R.~Ngo, H.~Noh, L.~Ouyang, C.~O'Keefe, J.~Pachocki, A.~Paino, J.~Palermo, A.~Pantuliano, G.~Parascandolo, J.~Parish, E.~Parparita,
  A.~Passos, M.~Pavlov, A.~Peng, A.~Perelman, F.~de~Avila Belbute~Peres, M.~Petrov, H.~P. de~Oliveira~Pinto, Michael, Pokorny, M.~Pokrass, V.~H. Pong, T.~Powell, A.~Power, B.~Power, E.~Proehl, R.~Puri, A.~Radford, J.~Rae, A.~Ramesh, C.~Raymond, F.~Real, K.~Rimbach, C.~Ross, B.~Rotsted, H.~Roussez, N.~Ryder, M.~Saltarelli, T.~Sanders, S.~Santurkar, G.~Sastry, H.~Schmidt, D.~Schnurr, J.~Schulman, D.~Selsam, K.~Sheppard, T.~Sherbakov, J.~Shieh, S.~Shoker, P.~Shyam, S.~Sidor, E.~Sigler, M.~Simens, J.~Sitkin, K.~Slama, I.~Sohl, B.~Sokolowsky, Y.~Song, N.~Staudacher, F.~P. Such, N.~Summers, I.~Sutskever, J.~Tang, N.~Tezak, M.~B. Thompson, P.~Tillet, A.~Tootoonchian, E.~Tseng, P.~Tuggle, N.~Turley, J.~Tworek, J.~F.~C. Uribe, A.~Vallone, A.~Vijayvergiya, C.~Voss, C.~Wainwright, J.~J. Wang, A.~Wang, B.~Wang, J.~Ward, J.~Wei, C.~Weinmann, A.~Welihinda, P.~Welinder, J.~Weng, L.~Weng, M.~Wiethoff, D.~Willner, C.~Winter, S.~Wolrich, H.~Wong, L.~Workman, S.~Wu, J.~Wu, M.~Wu, K.~Xiao, T.~Xu, S.~Yoo, K.~Yu, Q.~Yuan,
  W.~Zaremba, R.~Zellers, C.~Zhang, M.~Zhang, S.~Zhao, T.~Zheng, J.~Zhuang, W.~Zhuk, and B.~Zoph.
\newblock Gpt-4 technical report, 2024.
\newblock URL \url{https://arxiv.org/abs/2303.08774}.

\bibitem[Pan et~al.(2025)Pan, Liu, Wu, Liu, Zhu, Li, Chen, Ouyang, and Rueckert]{pan2025medvlmr1incentivizingmedicalreasoning}
J.~Pan, C.~Liu, J.~Wu, F.~Liu, J.~Zhu, H.~B. Li, C.~Chen, C.~Ouyang, and D.~Rueckert.
\newblock Medvlm-r1: Incentivizing medical reasoning capability of vision-language models (vlms) via reinforcement learning, 2025.
\newblock URL \url{https://arxiv.org/abs/2502.19634}.

\bibitem[Peng et~al.(2025)Peng, Chris, Wang, Wei, Pei, Qiu, Jian, Hao, Pan, Xie, Ge, Zhuang, Song, Liu, and Zhou]{peng2025skyworkr1vpioneeringmultimodal}
Y.~Peng, Chris, X.~Wang, Y.~Wei, J.~Pei, W.~Qiu, A.~Jian, Y.~Hao, J.~Pan, T.~Xie, L.~Ge, R.~Zhuang, X.~Song, Y.~Liu, and Y.~Zhou.
\newblock Skywork r1v: Pioneering multimodal reasoning with chain-of-thought, 2025.
\newblock URL \url{https://arxiv.org/abs/2504.05599}.

\bibitem[Simpson et~al.(2019)Simpson, Antonelli, Bakas, Bilello, Farahani, van Ginneken, Kopp-Schneider, Landman, Litjens, Menze, Ronneberger, Summers, Bilic, Christ, Do, Gollub, Golia-Pernicka, Heckers, Jarnagin, McHugo, Napel, Vorontsov, Maier-Hein, and Cardoso]{simpson2019largeannotatedmedicalimage}
A.~L. Simpson, M.~Antonelli, S.~Bakas, M.~Bilello, K.~Farahani, B.~van Ginneken, A.~Kopp-Schneider, B.~A. Landman, G.~Litjens, B.~Menze, O.~Ronneberger, R.~M. Summers, P.~Bilic, P.~F. Christ, R.~K.~G. Do, M.~Gollub, J.~Golia-Pernicka, S.~H. Heckers, W.~R. Jarnagin, M.~K. McHugo, S.~Napel, E.~Vorontsov, L.~Maier-Hein, and M.~J. Cardoso.
\newblock A large annotated medical image dataset for the development and evaluation of segmentation algorithms, 2019.
\newblock URL \url{https://arxiv.org/abs/1902.09063}.

\bibitem[Team et~al.(2025)Team, Du, Yin, Xing, Qu, Wang, Chen, Zhang, Du, Wei, Wang, Zhang, Du, Wang, Yuan, Lu, Li, Sung, Wei, Lai, Zhu, Ding, Hu, Yang, Zhang, Wu, Yao, Lu, Wang, Gao, Zheng, Li, Su, Wang, Deng, Qiu, Xie, Wang, Liu, Yan, Ouyang, Chen, Sui, Yu, Dong, Dong, Xu, Cheng, Gu, Zhou, Liu, Cao, Yu, Song, Bai, Song, He, Huang, Xu, Yuan, Yao, Wu, Zu, Zhou, Wang, Charles, Zhong, Li, Hu, Chen, Wang, Liu, Miao, Qin, Chen, Bao, Wang, Kang, Liu, Du, Wu, Wang, Yan, Zhou, Li, Jiang, Zhang, Yang, Huang, Huang, Zhao, Chen, and Lin]{kimiteam2025kimivltechnicalreport}
K.~Team, A.~Du, B.~Yin, B.~Xing, B.~Qu, B.~Wang, C.~Chen, C.~Zhang, C.~Du, C.~Wei, C.~Wang, D.~Zhang, D.~Du, D.~Wang, E.~Yuan, E.~Lu, F.~Li, F.~Sung, G.~Wei, G.~Lai, H.~Zhu, H.~Ding, H.~Hu, H.~Yang, H.~Zhang, H.~Wu, H.~Yao, H.~Lu, H.~Wang, H.~Gao, H.~Zheng, J.~Li, J.~Su, J.~Wang, J.~Deng, J.~Qiu, J.~Xie, J.~Wang, J.~Liu, J.~Yan, K.~Ouyang, L.~Chen, L.~Sui, L.~Yu, M.~Dong, M.~Dong, N.~Xu, P.~Cheng, Q.~Gu, R.~Zhou, S.~Liu, S.~Cao, T.~Yu, T.~Song, T.~Bai, W.~Song, W.~He, W.~Huang, W.~Xu, X.~Yuan, X.~Yao, X.~Wu, X.~Zu, X.~Zhou, X.~Wang, Y.~Charles, Y.~Zhong, Y.~Li, Y.~Hu, Y.~Chen, Y.~Wang, Y.~Liu, Y.~Miao, Y.~Qin, Y.~Chen, Y.~Bao, Y.~Wang, Y.~Kang, Y.~Liu, Y.~Du, Y.~Wu, Y.~Wang, Y.~Yan, Z.~Zhou, Z.~Li, Z.~Jiang, Z.~Zhang, Z.~Yang, Z.~Huang, Z.~Huang, Z.~Zhao, Z.~Chen, and Z.~Lin.
\newblock Kimi-vl technical report, 2025.
\newblock URL \url{https://arxiv.org/abs/2504.07491}.

\bibitem[Wang et~al.(2024{\natexlab{a}})Wang, Xu, Zhao, Ouyang, Wu, Zhao, Xu, Liu, Qu, Shang, Zhang, Wei, Sui, Li, Shi, Qiao, Lin, and He]{wang2024mineruopensourcesolutionprecise}
B.~Wang, C.~Xu, X.~Zhao, L.~Ouyang, F.~Wu, Z.~Zhao, R.~Xu, K.~Liu, Y.~Qu, F.~Shang, B.~Zhang, L.~Wei, Z.~Sui, W.~Li, B.~Shi, Y.~Qiao, D.~Lin, and C.~He.
\newblock Mineru: An open-source solution for precise document content extraction, 2024{\natexlab{a}}.
\newblock URL \url{https://arxiv.org/abs/2409.18839}.

\bibitem[Wang et~al.(2025{\natexlab{a}})Wang, Qu, Huang, Chu, Lin, and Chen]{vl-rethinker}
H.~Wang, C.~Qu, Z.~Huang, W.~Chu, F.~Lin, and W.~Chen.
\newblock Vl-rethinker: Incentivizing self-reflection of vision-language models with reinforcement learning.
\newblock \emph{arXiv preprint arXiv:2504.08837}, 2025{\natexlab{a}}.

\bibitem[Wang et~al.(2025{\natexlab{b}})Wang, Qu, Huang, Chu, Lin, and Chen]{wang2025vlrethinkerincentivizingselfreflectionvisionlanguage}
H.~Wang, C.~Qu, Z.~Huang, W.~Chu, F.~Lin, and W.~Chen.
\newblock Vl-rethinker: Incentivizing self-reflection of vision-language models with reinforcement learning, 2025{\natexlab{b}}.
\newblock URL \url{https://arxiv.org/abs/2504.08837}.

\bibitem[Wang et~al.(2024{\natexlab{b}})Wang, Pan, Shi, Lu, Ren, Zhou, Zhan, and Li]{wang2024measuring}
K.~Wang, J.~Pan, W.~Shi, Z.~Lu, H.~Ren, A.~Zhou, M.~Zhan, and H.~Li.
\newblock Measuring multimodal mathematical reasoning with math-vision dataset.
\newblock In \emph{The Thirty-eight Conference on Neural Information Processing Systems Datasets and Benchmarks Track}, 2024{\natexlab{b}}.
\newblock URL \url{https://openreview.net/forum?id=QWTCcxMpPA}.

\bibitem[Wang et~al.(2017)Wang, Peng, Lu, Lu, Bagheri, and Summers]{Wang_2017}
X.~Wang, Y.~Peng, L.~Lu, Z.~Lu, M.~Bagheri, and R.~M. Summers.
\newblock Chestx-ray8: Hospital-scale chest x-ray database and benchmarks on weakly-supervised classification and localization of common thorax diseases.
\newblock In \emph{2017 IEEE Conference on Computer Vision and Pattern Recognition (CVPR)}, page 3462–3471. IEEE, July 2017.
\newblock \doi{10.1109/cvpr.2017.369}.
\newblock URL \url{http://dx.doi.org/10.1109/CVPR.2017.369}.

\bibitem[Wang et~al.(2025{\natexlab{c}})Wang, Wu, Zhang, Yan, Liu, Luo, and Fei]{wang2025multimodalchainofthoughtreasoningcomprehensive}
Y.~Wang, S.~Wu, Y.~Zhang, S.~Yan, Z.~Liu, J.~Luo, and H.~Fei.
\newblock Multimodal chain-of-thought reasoning: A comprehensive survey, 2025{\natexlab{c}}.
\newblock URL \url{https://arxiv.org/abs/2503.12605}.

\bibitem[Wu et~al.(2023)Wu, Lin, Zhang, Zhang, Wang, and Xie]{wu2023pmcllamabuildingopensourcelanguage}
C.~Wu, W.~Lin, X.~Zhang, Y.~Zhang, Y.~Wang, and W.~Xie.
\newblock Pmc-llama: Towards building open-source language models for medicine, 2023.
\newblock URL \url{https://arxiv.org/abs/2304.14454}.

\bibitem[Wu et~al.(2024)Wu, Chen, Pan, Liu, Liu, Dai, Gao, Ma, Wu, Wang, Xie, Wu, Hu, Wang, Sun, Li, Piao, Guan, Liu, Xie, You, Dong, Yu, Zhang, Zhao, Wang, and Ruan]{wu2024deepseekvl2mixtureofexpertsvisionlanguagemodels}
Z.~Wu, X.~Chen, Z.~Pan, X.~Liu, W.~Liu, D.~Dai, H.~Gao, Y.~Ma, C.~Wu, B.~Wang, Z.~Xie, Y.~Wu, K.~Hu, J.~Wang, Y.~Sun, Y.~Li, Y.~Piao, K.~Guan, A.~Liu, X.~Xie, Y.~You, K.~Dong, X.~Yu, H.~Zhang, L.~Zhao, Y.~Wang, and C.~Ruan.
\newblock Deepseek-vl2: Mixture-of-experts vision-language models for advanced multimodal understanding, 2024.
\newblock URL \url{https://arxiv.org/abs/2412.10302}.

\bibitem[Yang et~al.(2024)Yang, Yang, Zhang, Hui, Zheng, Yu, Li, Liu, Huang, Wei, Lin, Yang, Tu, Zhang, Yang, Yang, Zhou, Lin, Dang, Lu, Bao, Yang, Yu, Li, Xue, Zhang, Zhu, Men, Lin, Li, Xia, Ren, Ren, Fan, Su, Zhang, Wan, Liu, Cui, Zhang, and Qiu]{qwen2.5}
A.~Yang, B.~Yang, B.~Zhang, B.~Hui, B.~Zheng, B.~Yu, C.~Li, D.~Liu, F.~Huang, H.~Wei, H.~Lin, J.~Yang, J.~Tu, J.~Zhang, J.~Yang, J.~Yang, J.~Zhou, J.~Lin, K.~Dang, K.~Lu, K.~Bao, K.~Yang, L.~Yu, M.~Li, M.~Xue, P.~Zhang, Q.~Zhu, R.~Men, R.~Lin, T.~Li, T.~Xia, X.~Ren, X.~Ren, Y.~Fan, Y.~Su, Y.~Zhang, Y.~Wan, Y.~Liu, Z.~Cui, Z.~Zhang, and Z.~Qiu.
\newblock Qwen2.5 technical report.
\newblock \emph{arXiv preprint arXiv:2412.15115}, 2024.

\bibitem[Yue et~al.(2024)Yue, Ni, Zhang, Zheng, Liu, Zhang, Stevens, Jiang, Ren, Sun, Wei, Yu, Yuan, Sun, Yin, Zheng, Yang, Liu, Huang, Sun, Su, and Chen]{yue2024mmmumassivemultidisciplinemultimodal}
X.~Yue, Y.~Ni, K.~Zhang, T.~Zheng, R.~Liu, G.~Zhang, S.~Stevens, D.~Jiang, W.~Ren, Y.~Sun, C.~Wei, B.~Yu, R.~Yuan, R.~Sun, M.~Yin, B.~Zheng, Z.~Yang, Y.~Liu, W.~Huang, H.~Sun, Y.~Su, and W.~Chen.
\newblock Mmmu: A massive multi-discipline multimodal understanding and reasoning benchmark for expert agi, 2024.
\newblock URL \url{https://arxiv.org/abs/2311.16502}.

\bibitem[Zhang et~al.(2024{\natexlab{a}})Zhang, Jiang, Zhang, Lin, Guo, Qiu, Zhou, Lu, Chang, Gao, and Li]{zhang2024mathversedoesmultimodalllm}
R.~Zhang, D.~Jiang, Y.~Zhang, H.~Lin, Z.~Guo, P.~Qiu, A.~Zhou, P.~Lu, K.-W. Chang, P.~Gao, and H.~Li.
\newblock Mathverse: Does your multi-modal llm truly see the diagrams in visual math problems?, 2024{\natexlab{a}}.
\newblock URL \url{https://arxiv.org/abs/2403.14624}.

\bibitem[Zhang et~al.(2024{\natexlab{b}})Zhang, Wu, Zhao, Lin, Zhang, Wang, and Xie]{zhang2024pmcvqavisualinstructiontuning}
X.~Zhang, C.~Wu, Z.~Zhao, W.~Lin, Y.~Zhang, Y.~Wang, and W.~Xie.
\newblock Pmc-vqa: Visual instruction tuning for medical visual question answering, 2024{\natexlab{b}}.
\newblock URL \url{https://arxiv.org/abs/2305.10415}.

\bibitem[Zhu et~al.(2025)Zhu, Wang, Chen, Liu, Ye, Gu, Tian, Duan, Su, Shao, Gao, Cui, Wang, Cao, Liu, Wei, Zhang, Wang, Xu, Li, Wang, Deng, Li, He, Jiang, Luo, Wang, He, Shi, Zhang, Shao, He, Xiong, Qu, Sun, Jiao, Lv, Wu, Zhang, Deng, Ge, Chen, Wang, Dou, Lu, Zhu, Lu, Lin, Qiao, Dai, and Wang]{zhu2025internvl3exploringadvancedtraining}
J.~Zhu, W.~Wang, Z.~Chen, Z.~Liu, S.~Ye, L.~Gu, H.~Tian, Y.~Duan, W.~Su, J.~Shao, Z.~Gao, E.~Cui, X.~Wang, Y.~Cao, Y.~Liu, X.~Wei, H.~Zhang, H.~Wang, W.~Xu, H.~Li, J.~Wang, N.~Deng, S.~Li, Y.~He, T.~Jiang, J.~Luo, Y.~Wang, C.~He, B.~Shi, X.~Zhang, W.~Shao, J.~He, Y.~Xiong, W.~Qu, P.~Sun, P.~Jiao, H.~Lv, L.~Wu, K.~Zhang, H.~Deng, J.~Ge, K.~Chen, L.~Wang, M.~Dou, L.~Lu, X.~Zhu, T.~Lu, D.~Lin, Y.~Qiao, J.~Dai, and W.~Wang.
\newblock Internvl3: Exploring advanced training and test-time recipes for open-source multimodal models, 2025.
\newblock URL \url{https://arxiv.org/abs/2504.10479}.

\end{thebibliography}
}


\newpage
\section*{NeurIPS Paper Checklist}

\begin{enumerate}

\item {\bf Claims}
    \item[] Question: Do the main claims made in the abstract and introduction accurately reflect the paper's contributions and scope?
    \item[] Answer: \answerYes{} 

\item {\bf Limitations}
    \item[] Question: Does the paper discuss the limitations of the work performed by the authors?
    \item[] Answer: \answerYes{} 
    \item[] Justification: See Section 6.

\item {\bf Theory assumptions and proofs}
    \item[] Question: For each theoretical result, does the paper provide the full set of assumptions and a complete (and correct) proof?
    \item[] Answer: \answerNA{} 
    \item[] Justification: Benchmark and dataset paper.

    \item {\bf Experimental result reproducibility}
    \item[] Question: Does the paper fully disclose all the information needed to reproduce the main experimental results of the paper to the extent that it affects the main claims and/or conclusions of the paper (regardless of whether the code and data are provided or not)?
    \item[] Answer: \answerNA{} 
    \item[] Justification: Benchmark and dataset paper.

\item {\bf Open access to data and code}
    \item[] Question: Does the paper provide open access to the data and code, with sufficient instructions to faithfully reproduce the main experimental results, as described in supplemental material?
    \item[] Answer: \answerYes{} 
    \item[] Justification: Important prompts used are available in the supplemental material. All our code and data are available to the general public. 

\item {\bf Experimental setting/details}
    \item[] Question: Does the paper specify all the training and test details (e.g., data splits, hyperparameters, how they were chosen, type of optimizer, etc.) necessary to understand the results?
    \item[] Answer: \answerYes{} 
    \item[] Justification: See section 5.

\item {\bf Experiment statistical significance}
    \item[] Question: Does the paper report error bars suitably and correctly defined or other appropriate information about the statistical significance of the experiments?
    \item[] Answer: \answerNo{} 
    \item[] Justification: We evaluated a large number of multimodal large language models. We did not conduct multiple runs of the experiments due to the high computational cost.

\item {\bf Experiments compute resources}
    \item[] Question: For each experiment, does the paper provide sufficient information on the computer resources (type of compute workers, memory, time of execution) needed to reproduce the experiments?
    \item[] Answer: \answerNo{} 
    \item[] Justification: Our experiments evaluated pre-trained multimodal large language models without any training or fine-tuning. The computing resource requirements for these models vary and are beyond our control.
    
\item {\bf Code of ethics}
    \item[] Question: Does the research conducted in the paper conform, in every respect, with the NeurIPS Code of Ethics \url{https://neurips.cc/public/EthicsGuidelines}?
    \item[] Answer: \answerYes{} 
    \item[] Justification: We have verified that the copyrights of the medical books from which our data is sourced are publicly available.

\item {\bf Broader impacts}
    \item[] Question: Does the paper discuss both potential positive societal impacts and negative societal impacts of the work performed?
    \item[] Answer: \answerYes{} 
    \item[] Justification: See sections 6 and 7.

\item {\bf Safeguards}
    \item[] Question: Does the paper describe safeguards that have been put in place for responsible release of data or models that have a high risk for misuse (e.g., pretrained language models, image generators, or scraped datasets)?
    \item[] Answer: \answerNA{} 

\item {\bf Licenses for existing assets}
    \item[] Question: Are the creators or original owners of assets (e.g., code, data, models), used in the paper, properly credited and are the license and terms of use explicitly mentioned and properly respected?
    \item[] Answer: \answerYes{} 
    \item[] Justification: We have verified that the copyrights of the medical books from which our data is sourced are publicly available. We use license CC BY 4.0.

\item {\bf New assets}
    \item[] Question: Are new assets introduced in the paper well documented and is the documentation provided alongside the assets?
    \item[] Answer: \answerYes{} 
    \item[] Justification: We include documentation when we release our dataset.

\item {\bf Crowdsourcing and research with human subjects}
    \item[] Question: For crowdsourcing experiments and research with human subjects, does the paper include the full text of instructions given to participants and screenshots, if applicable, as well as details about compensation (if any)? 
    \item[] Answer: \answerNA{} 

\item {\bf Institutional review board (IRB) approvals or equivalent for research with human subjects}
    \item[] Question: Does the paper describe potential risks incurred by study participants, whether such risks were disclosed to the subjects, and whether Institutional Review Board (IRB) approvals (or an equivalent approval/review based on the requirements of your country or institution) were obtained?
    \item[] Answer: \answerNA{} 

\item {\bf Declaration of LLM usage}
    \item[] Question: Does the paper describe the usage of LLMs if it is an important, original, or non-standard component of the core methods in this research? Note that if the LLM is used only for writing, editing, or formatting purposes and does not impact the core methodology, scientific rigorousness, or originality of the research, declaration is not required.
    \item[] Answer: \answerYes{} 
    \item[] Justification: See section 3, where we describe how we used LLM in our method. We also include important prompts in suplimentary materials.

\end{enumerate}

\newpage
\section*{Appendix}
\addcontentsline{toc}{section}{Appendix}

\begin{center}
\end{center}
\begin{enumerate}
    \item \hyperref[sec:query_keywords]{List of Query Keywords}
    \item \hyperref[sec:data_collection]{Prompts Used in Data Collection}
    \begin{enumerate}
        \item \hyperref[sec:categorization_prompts]{Prompts for Categorization}
        \item \hyperref[sec:figtexts_prompt]{Prompt for \textit{FIGtexts} Collection}
    \end{enumerate}
    \item \hyperref[sec:benchmark_construction]{Prompts Used in Benchmark Construction}
    \begin{enumerate}
        \item \hyperref[sec:vqa_generation]{Prompt for VQA Generation}
        \item \hyperref[sec:vqa_to_mc]{Prompt for Reformatting VQA to MC}
        \item \hyperref[sec:suitability_filtering]{Prompts for \textit{Suitability Filtering} and  \textit{Multimodality Filtering}}
        \item \hyperref[sec:labeling_benchmark]{Prompts for Labeling Benchmark}
    \end{enumerate}
    \item \hyperref[sec:distribution]{Distribution by Modality, Department, and Anatomy}
    \begin{enumerate}
        \item \hyperref[sec:modality_distribution]{Modality}
        \item \hyperref[sec:department_distribution]{Department}
        \item \hyperref[sec:anatomy_distribution]{Anatomy}
    \end{enumerate}
    \item \hyperref[sec:results_table]{Table of Results}
    \item \hyperref[sec:sample_entries]{Sample Entries in MedBookVQA}
\end{enumerate}

\appendix

\vspace{80pt}
\section{List of Query Keywords}
\label{sec:query_keywords}
The data for this study were sourced from medical books downloaded from the Directory of Open Access Books (DOAB). To effectively collect relevant books, we utilized the website's keyword search feature with a comprehensive list of medical terms. These keywords, which span various medical specialties and imaging techniques, are presented in Table~\ref{tab:keywords}.

\begin{table}[ht] 
\centering 
\caption{Medical keywords used to query DOAB for book collection.} 
\label{tab:keywords} 
\resizebox{\textwidth}{!}{
\begin{tabular}{|l|}  
\hline 
X-ray, MRI, Magnetic Resonance Imaging, CT, Computed Tomography, Ultrasound, OCT,  \\ 
Optical Coherence Tomography, Positron Emission Tomography Electrocardiogram, Colposcopy, \\ 
Dermoscopy, Endoscopy, Fundus Photography, Microscopy Images, Infrared Reflectance Imaging,  \\
Fluid-Attenuated Inversion Recovery, Pathology, Radiology, Dermatology, Neurology, Anatomy,  \\ 
Otolaryngology, Orthopedics, Gastroenterology, Hematology, Obstetrics and Gynecology, \\ 
Ophthalmology, Physiology, Psychiatry, Pediatrics, Pharmacology, Surgery, Dentistry,  \\ 
Orthodontics, Anesthesiology, Cardiology, Oncology, Immunology, Endocrinology, Urology,  \\ 
Nephrology, Pulmonology, Biochemistry, Infectious Diseases, Medical, Nursing, Neuroscience, \\ 
Critical Care Medicine, Nutrition, Emergency Medicine, Cell Biology and Histology,  \\ 
Pain Medicine, Physical Medicine and Rehabilitation \\ 
\hline 
\end{tabular} 
}
\end{table}

\newpage
\section{Prompts Used in Data Collection}
\label{sec:data_collection}
This section outlines the prompts utilized in the pipeline to process medical books into figure-information pairs and generate Visual Question Answering (VQA) questions. Figure \ref{prompt_categorization} illustrates the prompt designed to categorize figures into four distinct categories. Additionally, Figure \ref{prompt_FIGtexts} presents the prompts used to collect sentences that describe figures by explicitly referencing the figure's name. These prompts ensure that the descriptions are precise and directly tied to the visual content.

\subsection{Prompt for Categorization}
\label{sec:categorization_prompts}

\begin{figure}[!ht]
\begin{promptenv}{Definition of Categories}
\textbf{Category "A": real-case medical figure.} \\
Medical figures gathered in real-world scenarios.  \\
(common types: photos taken of human bodies, especially disease-related areas;
    photos taken using medical equipment, such as: X-ray, MRI, CT, Ultrasound, OCT, PET, Electrocardiogram,
    Histopathology, Colposcopy, Dermoscopy, Endoscopy, Fundus Photography, Microscopy Images, Digital
    Photography, Infrared Reflectance Imaging, Fluid-Attenuated Inversion Recovery, DTI.) \\

\textbf{Category "B": Conceptual illustration of medical knowledge.}  \\
(common types: abstract non-realistic illustrations for human bodies; graphs, charts, or tables to explain or arrange medical knowledge information;
    photos of medical or laboratory equipment; illustrations of how to use medical or laboratory equipment;
    illustrations of interview-like experiment settings.) \\

\textbf{Category "C": Charts of data results from medical research or experiments.} \\
(common types: data charts, tables, graphs, etc., to summarize data results) \\

\textbf{Category "D": others.} \\
(Any other types of figures not included in the above categories, such as logos, icons, famous specialists, etc.) \\

\end{promptenv}
\begin{promptenv}{Prompt for Figure Categorization.}
The image above is a figure in a medical resource with reference name: \{FIGname\}. \\
The following is the caption of the figure: \\
<caption starts> \\
\{caption\} \\
<caption ends> \\

According to the figure and with the help of the caption, categorize the figure into one of the following categories: \\
<category definition start> \\
\{DefCategory\} \\
<category definition end> \\

Treat the figure as the target, while the caption and texts as assistant references only. \\

Response ONLY the category letter: (A, B, C, D).  \\
If you do not know the answer, just give the letter "D".
\end{promptenv}
\caption{Prompt used to categorize figures in books into four categories via two methods: \textit{FifOnly} and \textit{Fig\&Caption}.}
\label{prompt_categorization}
\end{figure}

\newpage
\subsection{Prompt for \textit{FIGtexts} Collection}
\label{sec:figtexts_prompt}

\begin{figure}[!ht]
\begin{promptenv}{Prompt for FIGtexts Collection.}
The image above is a figure in a medical resource with reference name: \{FIGname\}. \\
The following is a text paragraph near the figure: \\
<paragraph starts> \\
\{paragraph\} \\
<paragraph ends> \\

Find all complete sentences describe "\{FIGname\}" in the paragraph (information only between <paragraph starts> and <paragraph ends>) given above,
by explicitly mentioning the word "\{FIGname\}" in the sentences. \\
Note that there may be other figures or tables mentioned in the paragraph. Do not mess with the one focused on. \\

Response ONLY in the following specific format (@\#\$ will be used for separation): \\
\{ \\
@\#\$ <piece of sentences> \\
@\#\$ <piece of sentences> \\
@\#\$ ... \\
\}
\end{promptenv}
\caption{Prompt used to collect \textit{FIGtexts}, which are sentences that describe figures by explicitly mentioning the figure reference name.}
\label{prompt_FIGtexts}
\end{figure}

\newpage
\section{Prompt Used in Benchmark Construction}
\label{sec:benchmark_construction}
This section details the prompts employed in the pipeline to construct the MedBookVQA benchmark from the processed figure-information pairs. The pipeline generates five types of VQA questions (as shown in Figure \ref{prompt_GenVQA}) and reformats them into multiple-choice (MC) format (Figure \ref{prompt_VQAtoMC}). The questions are then filtered based on their suitability and whether they require multimodal information to be answered (Figures \ref{Prompt_suitabilityFilter} and \ref{prompt_MMfilter}). Finally, the data are annotated with labels categorizing them by modality, anatomy, and department, as demonstrated in Figures \ref{prompt_Distribution_anatomy}, \ref{prompt_Distribution_modality}, and \ref{prompt_Distribution_department}.

\subsection{Prompt for VQA Generation}
\label{sec:vqa_generation}

\begin{figure}[!ht]
\begin{promptenv}{Definition of VQA Types.}
\textbf{(M): Modality Recognition}: Identify the modality of the image(e.g. MRI, CT, X-ray, etc.) \\
\textbf{(D): Disease Recognition}: Identify the disease/condition shown in the image. \\
\textbf{(A): Anatomy Identification}: Identify the anatomy / body structure / organ shown in the image. \\
\textbf{(S): Symptom Diagnosis}: Determine the symptom / abnormality diagnosis / abnormality description / abnormality position shown in the image. \\
\textbf{(SO): Surgery \& Operation}: (Consider only if it is a surgical/operational image). Prepare VQA related to significant information such as surgeon actions/operations, instructions, surgical techniques, steps/workflows, tool/artifact/instrument usage, positional arrangement of the patient, physical treatments, etc., in the surgery / medical operation. \\
\end{promptenv}

\begin{promptenv}{Prompt for VQA Generation.}
Generate at most 6 medical question-answer(QA) pairs about the figure. \\
Requirements:  \\
1. QA pair should be one of the following QAtypes (ONLY choose APPLICABLE ones):  \\
<type definition starts> \\
\{QAtypes\} \\
<type definition ends> \\
2. Please refer to the following details of each QAtypes: \\
<type details starts> \\
\{QAtype\_details\} \\
<type details ends> \\
3. For each pair provide: QAtpyeID(e.g. M), question, answer. \\
4. Keep the "answer" as CONCISE as possible. Try to use several words or a short phrase. \\
5. Only based on the information given above;  \\
6. Do NOT include any identifiers(i.e., figure name) in the generated QA pairs. \\
7. Each question should be answerable for the case when ONLY the figure is given,  \\
which means do NOT ask the answerer to refer to any textual information in the question. \\
8. Do NOT assume the answerer can access the textual information. \\
9. If there are multiple images, clearly specify which image the QA pair is referring to in the question part. \\

Response ONLY in the following specific format (@\#\$ will be used for separation): \\
\{ \\
@@\#\#\$\$ <QAtpyeID> @\#\$ <question> @\#\$ <answer> \\
@@\#\#\$\$ ... \\
\}  \\
\end{promptenv}

\caption{Prompt used to generate VQAs from a figure-information pair.}
\label{prompt_GenVQA}
\end{figure}

\newpage
\subsection{Prompt for Reformatting VQA to MC}
\label{sec:vqa_to_mc}

\begin{figure}[!ht]
\begin{promptenv}{Prompt for Reformatting VQA to MC.}
You are a medical specialist with expertise in medical knowledge across various diseases. \\
Reformat the pair into a multiple-choice question by generating three distractors (incorrect answer choices). \\
Requirements: \\
(1) Consider the original answer as the correct answer choice. \\
(2) Generate three distractors for the multiple-choice question. \\
(3) Ensure the distractors have a similar length and sentence structure to the correct answer. \\
(4) Ensure the distractors are incorrect and challenging. \\
(5) Avoid using right/left distinctions to generate distractors. For example, if "right lung" appears in the answer, do not generate a distractor by replacing it with "left lung". \\
(6) Avoid using superordinate category terms to generate distractors. For example, if "T1 MRI" appears in the answer, do not generate "MRI" as a distractor. \\
(7) For "Anatomy Identification" type, do not use anatomical structures that also appear in the image to generate distractors. \\
(8) Only respond with the distractors in the following format, using @\#\$ as a separator:
"@\#\$ distractor1 @\#\$ distractor2 @\#\$ distractor3" \\
\end{promptenv}
\caption{Prompt used to reformat a VQA into MC.}
\label{prompt_VQAtoMC}
\end{figure}

\subsection{Prompt for \textit{Suitability Filtering} and \textit{Multimodality Filtering}}
\label{sec:suitability_filtering}
\begin{figure}[!ht]
\begin{promptenv}{Prompt for \textit{Suitability Filtering}.}
You are a medical specialist with expertise in medical knowledge across various diseases. \\
Based on the image, image-related description, determine if the image is about real-case medical scenarios of humans for disease diagnostic purposes or surgical operation-related. It should meet the following requirements: \\
(1) common types: photos taken for human body, especially disease related area; photos taken using medical equipment, such as: X-ray, MRI, CT, Ultrasound, OCT, PET, Electrocardiogram, Histopathology, Colposcopy, Dermoscopy, Endoscopy, Fundus Photography, Microscopy Images, Digital Photography, Infrared Reflectance Imaging, Fluid-Attenuated Inversion Recovery, DTI.) \\
(2) Determine "real-case" based on the image itself, not the sources, which means an image from another publication is also acceptable. \\
(3) Images from other subjects should be excluded, e.g., biology, microbiology, plants, animals, veterinary medicine, forensic medicine, etc. \\
(4) If suitable, respond with "Suitable.". If not, respond with "Not suitable. <brief reason>.".
\end{promptenv}
\caption{Prompt used to filter VQA based on suitability.}
\label{Prompt_suitabilityFilter}
\end{figure}

\begin{figure}[!ht]
\begin{promptenv}{Prompt for \textit{Multimodality Filtering}.}
You are a medical specialist with expertise in medical knowledge across various diseases. \\
Your task is to determine if the following question-answering pair (QA) is answerable with the current information or requires additional context to answer. \\
If the QA pair is answerable with the current information, respond with "ANSWERABLE"; otherwise, respond with "NON-ANSWERABLE". \\
\end{promptenv}
\caption{Prompt used to filter VQA based on whether they require multimodal elements to answer. Note that this step uses LLM.}
\label{prompt_MMfilter}
\end{figure}

\newpage
\subsection{Prompt for Labeling Benchmark}
\label{sec:labeling_benchmark}

\begin{figure}[!ht]
\begin{promptenv}{Prompt for Anatomy.}
Identify the anatomy of the image with the help of the description and question-answer pair (QA). \\
<Description Start> \\
\{description\} \\
<Description End> \\
<QA Start> \\
Q: \{question\} \\
A: \{answer\} \\
<QA End> \\

Choose one of the following options based on the image and description. If more than one is suitable, choose the most relevant one under the QA context: \\
<Anatomy Options Start> \\
\{ANATOMY\_Organ\_list\} \# ANATOMY\_bodyPart\_list  \\
<Anatomy Options End> \\
Or respond with "Other" if you can not determine.  \\

Do not explain anything or the details.  \\
Respond with ONLY one of the "Anatomy Options" or respond ONLY with the word "Other". \\
\end{promptenv}
\caption{Prompt used to label an entry in the benchmark according to anatomy.}
\label{prompt_Distribution_anatomy}
\end{figure}

\begin{figure}[!ht]
\begin{promptenv}{Prompt for Modality.}
Identify the modality of the image with the help of the description. \\
<Description Start> \\
\{description\} \\
<Description End> \\

Choose one of the following options (the one in the bracket is the abbreviation) if mentioned in the description: \\
<Modality Options Start> \\
\{MODALITY\_list\} \\
<Modality Options End> \\
If none of the above "Modality Options" is suitable and the image is one type of medical diagnosis imaging, determine the modality by yourself. \\
If the image is not one type of medical diagnosis imaging, please determine if it is of type: "General Photo of Affected Area", which are photos taken by general photography tools, such as a camera or smartphone, that show the affected area of the body. \\
If all the above cases fail, please respond "Other". \\

Do not explain anything or the details.  \\
Respond with ONLY one of the "Modality Options" or ONLY the phrase "General Photo of Affected Area" or ONLY the word "Other". \\
\end{promptenv}
\caption{Prompt used to label an entry in the benchmark according to modality.}
\label{prompt_Distribution_modality}
\end{figure}

\begin{figure}[!ht]
\begin{promptenv}{Prompt for Department.}
Identify the medical specialty of the image with the help of the description and question-answer pair (QA). \\
<Description Start> \\
\{description\} \\
<Description End> \\
<QA Start> \\
Q: \{question\} \\
A: \{answer\} \\
<QA End> \\

Choose one of the following options based on the image and description. If more than one are suitable, choose the most relevant one under the QA context: \\
<Specialty Options Start> \\
\{DEPARTMENT\_list\} \\
<Specialty Options End> \\
If none of the above "Specialty Options" can be identified, you may determine the Specialty by yourself.
If you cannot determine, respond with "Other".  \\

Do not explain anything or the details.  \\
Respond with ONLY one of the "Specialty Options" or ONLY with the word "Other". \\
\end{promptenv}
\caption{Prompt used to label an entry in the benchmark according to department.}
\label{prompt_Distribution_department}
\end{figure}

\clearpage
\newpage
\section{Distribution by Modality, Department, and Anatomy}
\label{sec:distribution}
\begin{figure}[!th]
\centering
\vspace{-20pt}
\includegraphics[width=1\textwidth]{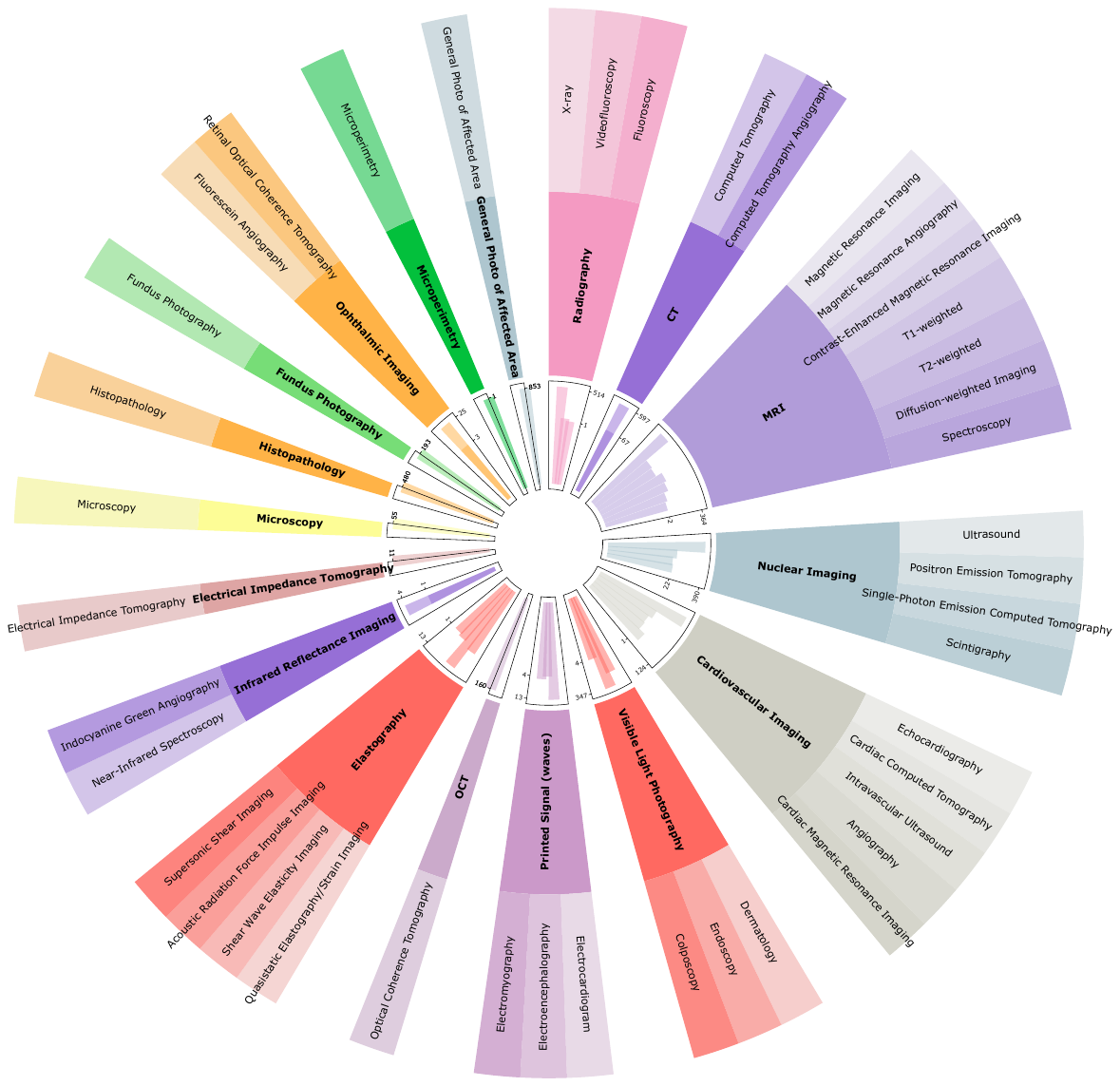}
\caption{Distribution in terms of modality, with \textit{ModalityCategory} as the inner level and \textit{Modality} as the outer level. The innermost bar charts show the amount of each \textit{Modality}, with minimum and maximum values displayed.}
\label{Modality_distribution}
\end{figure}

\begin{figure}[!th]
\centering
\includegraphics[width=1\textwidth]{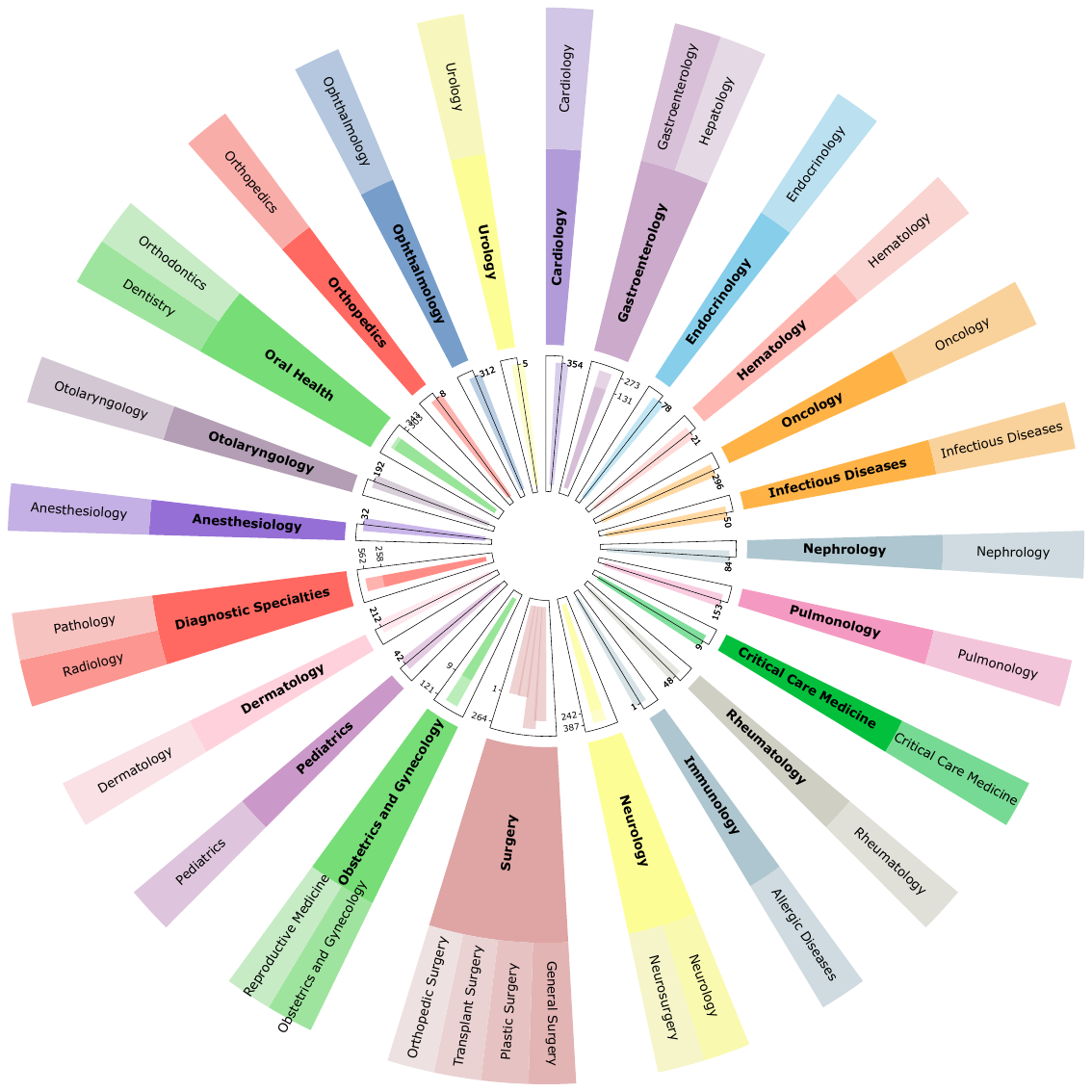}
\caption{Distribution in terms of department, structured with \textit{DepartmentCategory} as the inner level and \textit{Department} as the outer level. The innermost bar charts show the amount of each \textit{Department}, with minimum and maximum values displayed.}
\label{Department_distribution}
\end{figure}

\begin{figure}[!th]
\centering
\includegraphics[width=1\textwidth]{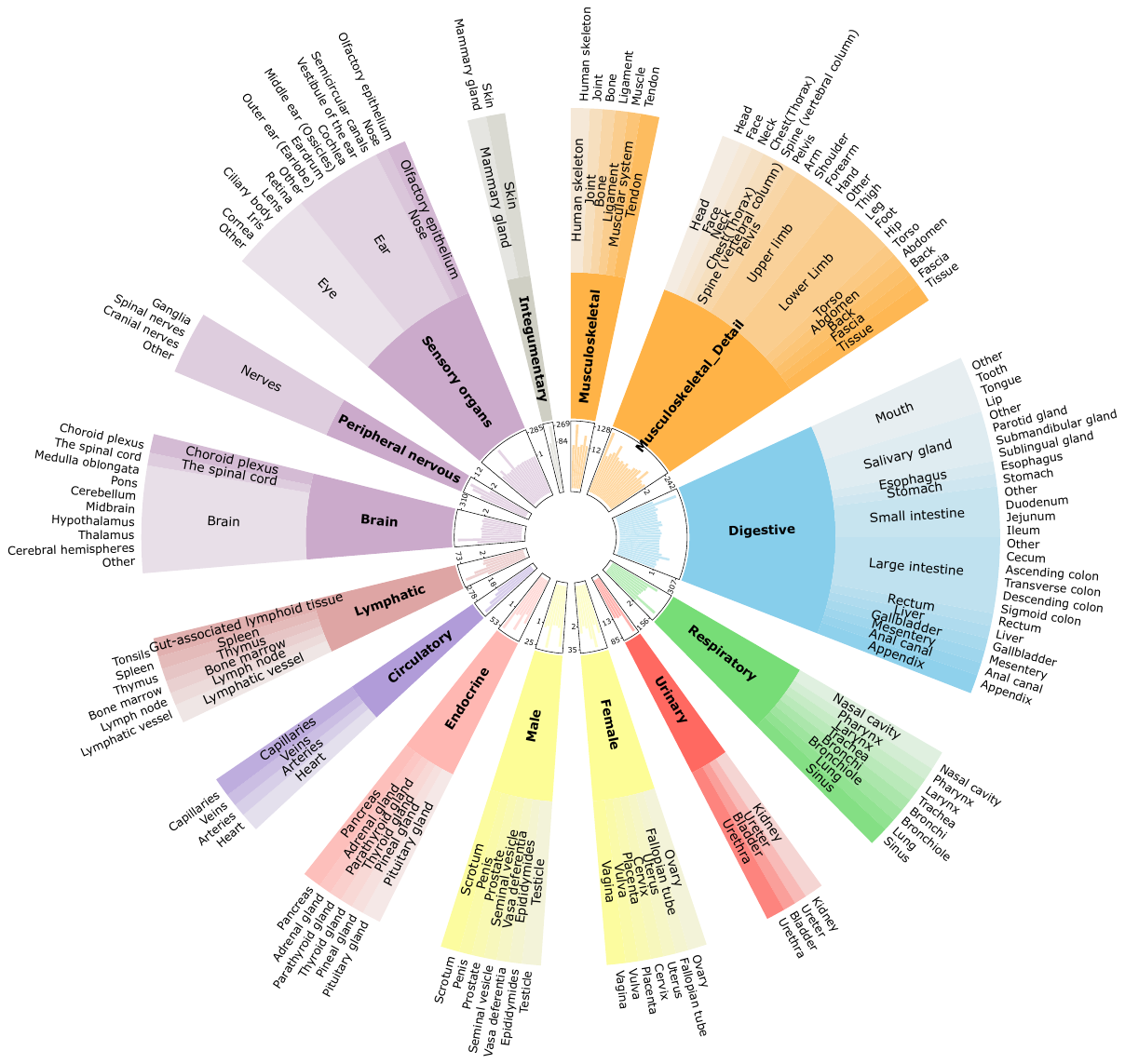}
\caption{Distribution in terms of anatomy. The levels, from inner to outer, are: \textit{SubSystem}, \textit{BodyPart}, and \textit{Organ}. The highest level, \textbf{\textit{System}}, is not shown. The relationships between \textit{System} and \textit{SubSystem} are as follows: \textit{Musculoskeletal\_Detail} and \textit{Musculoskeletal} belong to the \textit{\textbf{Musculoskeletal}} system; \textit{Female} and \textit{Male} belong to the \textit{\textbf{Reproductive}} system; \textit{Circulatory} and \textit{Lymphatic} belong to the \textit{\textbf{Circulatory}} system; \textit{Brain}, \textit{Peripheral nervous}, and \textit{Sensory organs} belong to the \textit{\textbf{Nervous}} system; others \textit{SubSystem} share the same names with their \textit{System}. The innermost bar charts show the amount of each \textit{Organ}, with minimum and maximum values displayed.}
\label{Anatomy_distribution}
\end{figure}

This section details the distribution of MedBookVQA across three key dimensions: anatomy, modality, and department, utilizing a hierarchical label system. The modality and department each employ a two-level structure, while anatomy is categorized using a four-level tree structure, as illustrated in Figures \ref{Modality_distribution}, \ref{Department_distribution}, and \ref{Anatomy_distribution}.

\subsection{Modality}
\label{sec:modality_distribution}

\newpage
\subsection{Department}
\label{sec:department_distribution}

\newpage
\subsection{Anatomy}
\label{sec:anatomy_distribution}

\newpage
\section{Table of Results}
\label{sec:results_table}
This section presents the evaluation results in detail. Performance is measured as the percentage of correctly answered questions across all categories and each of the five VQA types, with each type comprising 1,000 questions (totaling 5,000 questions). The models are categorized into four groups: \textbf{Proprietary General MLLMs}, \textbf{Open-Sourced General MLLMs}, \textbf{Open-Sourced Medical MLLMs}, and \textbf{Reasoning MLLMs}. Within and across categories, models are sorted by performance in descending order, with models from the same series further ordered by size.

\vspace{30pt}
\begin{table}[!th]
\caption{Evaluation results of all models, measured as the percentage of correctly answered questions across different categories.}
\label{tab:icl_exp}
\centering
\resizebox{1.0\textwidth}{!}{
\begin{tabular}{lcccccc}
\toprule
\textbf{Model} & \textbf{ALL} & \parbox{2.2cm}{\centering \textbf{Modality} \\ \textbf{Recognition}} & \parbox{2.2cm}{\centering \textbf{Disease} \\ \textbf{Recognition}} & \parbox{2.2cm}{\centering \textbf{Anatomy} \\ \textbf{Identification}} & \parbox{2.2cm}{\centering \textbf{Surgery \&} \\ \textbf{Operation}} & \parbox{2.2cm}{\centering \textbf{Symptom} \\ \textbf{Diagnosis}} \\
\midrule
\rowcolor{orange!30}\multicolumn{7}{c}{\textbf{\textit{Proprietary General MLLMs}}} \\
Gemini2.5-Pro (2025-03-26) & 81.24 & 95.60 & 72.90 & 82.80 & 77.80 & 77.10 \\
GPT4.1 & 77.98 & 94.60 & 71.30 & 77.00 & 74.50 & 72.50 \\
GPT4.1-mini & 75.02 & 92.50 & 64.70 & 73.70 & 71.20 & 73.00 \\
GPT-4o & 74.94 & 93.20 & 67.10 & 74.20 & 70.60 & 69.60 \\
Claude3.7-Sonnet & 67.72 & 88.60 & 55.00 & 67.10 & 63.40 & 64.50 \\
\midrule
\rowcolor{orange!30}\multicolumn{7}{c}{\textbf{\textit{Open-Sourced General MLLMs}}} \\
InternVL3-78B & 72.92 & 90.30 & 61.60 & 69.00 & 71.80 & 71.90 \\
InternVL3-38B & 70.44 & 89.40 & 59.70 & 70.80 & 66.90 & 65.40 \\
InternVL3-14B & 66.04 & 85.50 & 55.30 & 66.10 & 61.10 & 62.20 \\
InternVL3-9B & 64.42 & 85.60 & 52.90 & 63.50 & 58.70 & 61.40 \\
InternVL3-8B & 64.06 & 86.00 & 52.40 & 65.00 & 56.50 & 60.40 \\
InternVL3-2B & 59.12 & 83.00 & 45.60 & 63.40 & 51.70 & 51.90 \\
InternVL3-1B & 49.28 & 72.90 & 36.30 & 56.70 & 39.00 & 41.50 \\
LLaVA-OV-72B & 72.86 & 89.30 & 63.30 & 69.60 & 72.10 & 70.00 \\
LLaVA-OV-7B & 60.96 & 84.40 & 50.20 & 60.40 & 53.20 & 56.60 \\
LLaVA-OV-0.5B & 41.40 & 59.70 & 33.70 & 45.20 & 34.80 & 33.60 \\
Ovis2-34B & 71.42 & 88.20 & 62.20 & 70.90 & 67.70 & 68.10 \\
Ovis2-16B & 68.90 & 87.30 & 60.00 & 67.20 & 64.20 & 65.80 \\
Ovis2-8B & 68.12 & 87.50 & 57.70 & 66.90 & 62.40 & 66.10 \\
Ovis2-4B & 63.30 & 85.20 & 53.20 & 66.00 & 54.10 & 58.00 \\
Ovis2-2B & 62.00 & 81.70 & 49.10 & 64.40 & 56.10 & 58.70 \\
Ovis2-1B & 48.58 & 75.50 & 34.90 & 53.30 & 37.70 & 41.50 \\
Qwen2.5-VL-72B-Instruct & 70.06 & 89.90 & 60.20 & 70.70 & 63.00 & 66.50 \\
Qwen2.5-VL-32B-Instruct & 66.46 & 88.10 & 57.20 & 67.70 & 60.00 & 59.30 \\
Qwen2.5-VL-7B-Instruct & 63.44 & 87.50 & 49.80 & 67.70 & 55.00 & 57.20 \\
Qwen2.5-VL-3B-Instruct & 57.82 & 84.20 & 42.90 & 60.80 & 51.00 & 50.20 \\
InternVL2.5-78B  & 69.26 & 89.80 & 56.90 & 67.90 & 65.30 & 66.40 \\
InternVL2.5-38B  & 67.18 & 86.80 & 55.50 & 66.90 & 63.60 & 63.10 \\
InternVL2.5-26B  & 62.30 & 85.80 & 49.50 & 65.60 & 54.30 & 56.30 \\
InternVL2.5-8B & 60.90 & 82.00 & 52.30 & 62.80 & 52.10 & 55.30 \\
InternVL2.5-4B & 58.68 & 82.30 & 44.90 & 60.50 & 52.00 & 53.70 \\
InternVL2.5-2B & 47.70 & 70.80 & 36.00 & 53.00 & 39.30 & 39.40 \\
InternVL2.5-1B  & 46.90 & 72.70 & 32.60 & 53.50 & 38.30 & 37.40 \\
DeepSeek-VL2-Tiny & 44.60 & 76.20 & 33.30 & 47.20 & 33.70 & 32.60 \\
DeepSeek-VL2-Small & 34.94 & 52.50 & 29.00 & 36.60 & 30.60 & 26.00 \\
\midrule
\rowcolor{orange!30}\multicolumn{7}{c}{\textbf{\textit{Open-Sourced Medical MLLMs}}} \\
HealthGPT-L14 & 68.58 & 85.00 & 63.00 & 66.00 & 64.50 & 64.40 \\
HealthGPT-M3 & 64.52 & 80.60 & 59.50 & 64.50 & 57.20 & 60.80 \\
HuatuoGPT-Vision-34B & 66.10 & 89.70 & 55.50 & 69.10 & 56.80 & 59.40 \\
HuatuoGPT-Vision-7B & 62.82 & 88.40 & 53.50 & 67.30 & 51.00 & 53.90 \\
\midrule
\rowcolor{orange!30}\multicolumn{7}{c}{\textbf{\textit{Reasoning MLLMs}}} \\
\rowcolor{black!30}
\multicolumn{1}{c}{\textbf{Proprietary General Reasoning MLLMs}} \\
Claude3.7-Sonnet-Thinking & 67.50 & 89.60 & 57.10 & 66.70 & 60.90 & 63.20 \\
\midrule
\rowcolor{black!30}
\multicolumn{1}{c}{\textbf{Open-Sourced General Reasoning MLLMs}} \\
VL-Reasoner-72B & 61.40 & 85.00 & 50.20 & 63.10 & 54.10 & 54.60 \\
Skywork-R1V2-38B & 64.70 & 82.00 & 56.30 & 63.20 & 61.00 & 61.00 \\
Skywork-R1V-38B & 56.20 & 76.30 & 44.60 & 58.20 & 50.20 & 51.70 \\
Kimi-VL-A3B-Thinking  & 60.80 & 82.40 & 50.90 & 61.90 & 54.90 & 53.90 \\
\midrule
\rowcolor{black!30}
\multicolumn{1}{c}{\textbf{Open-Sourced Medical Reasoning MLLMs}} \\
MedVLM-R1 & 57.68 & 83.00 & 41.80 & 62.20 & 53.70 & 47.70 \\
\midrule
\textbf{Total MCQ} & \textbf{5,000} & \textbf{1,000} & \textbf{1,000} & \textbf{1,000} & \textbf{1,000} & \textbf{1,000} \\
\bottomrule
\end{tabular}
}
\end{table}

\newpage
\section{Sample Entries in MedBookVQA}
\label{sec:sample_entries}

This section provides examples selected from five VQA types, focusing on the most prevalent medical aspects (those with over 200 entries). For each entry, the hierarchical labels, VQA types, question, answer choices, and the correct answer are presented. Additionally, responses from the high-performing models across each of the four categories are included.

\begin{figure}[!th] 
\vspace{80pt}
\centering
\includegraphics[width=1\textwidth]{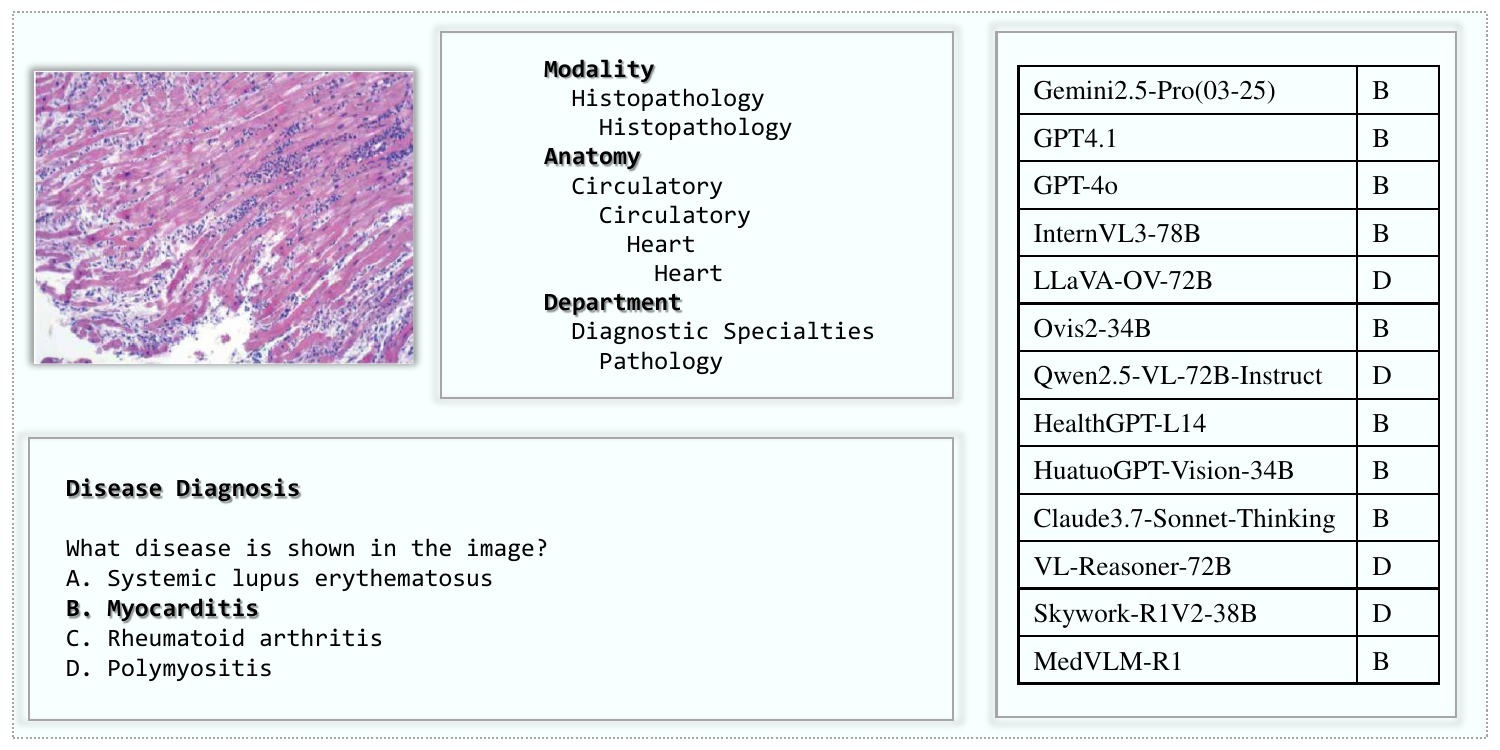}
\label{Entry_BodyPart_Heart}
\end{figure}

\begin{figure}[!th] 
\centering
\includegraphics[width=1\textwidth]{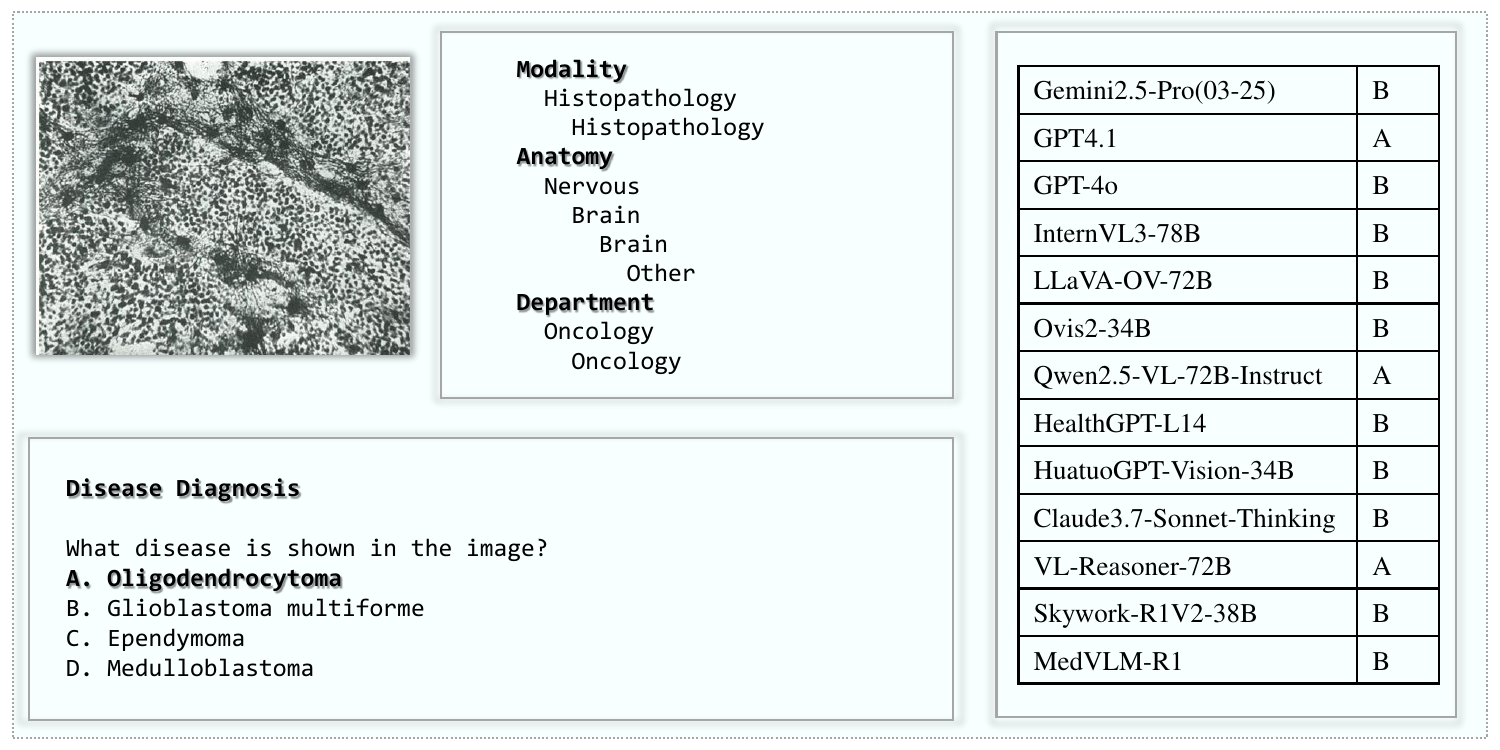}
\label{Entry_BodyPart_Brain}
\end{figure}

\newpage
\begin{figure}[!th] 
\centering
\includegraphics[width=1\textwidth]{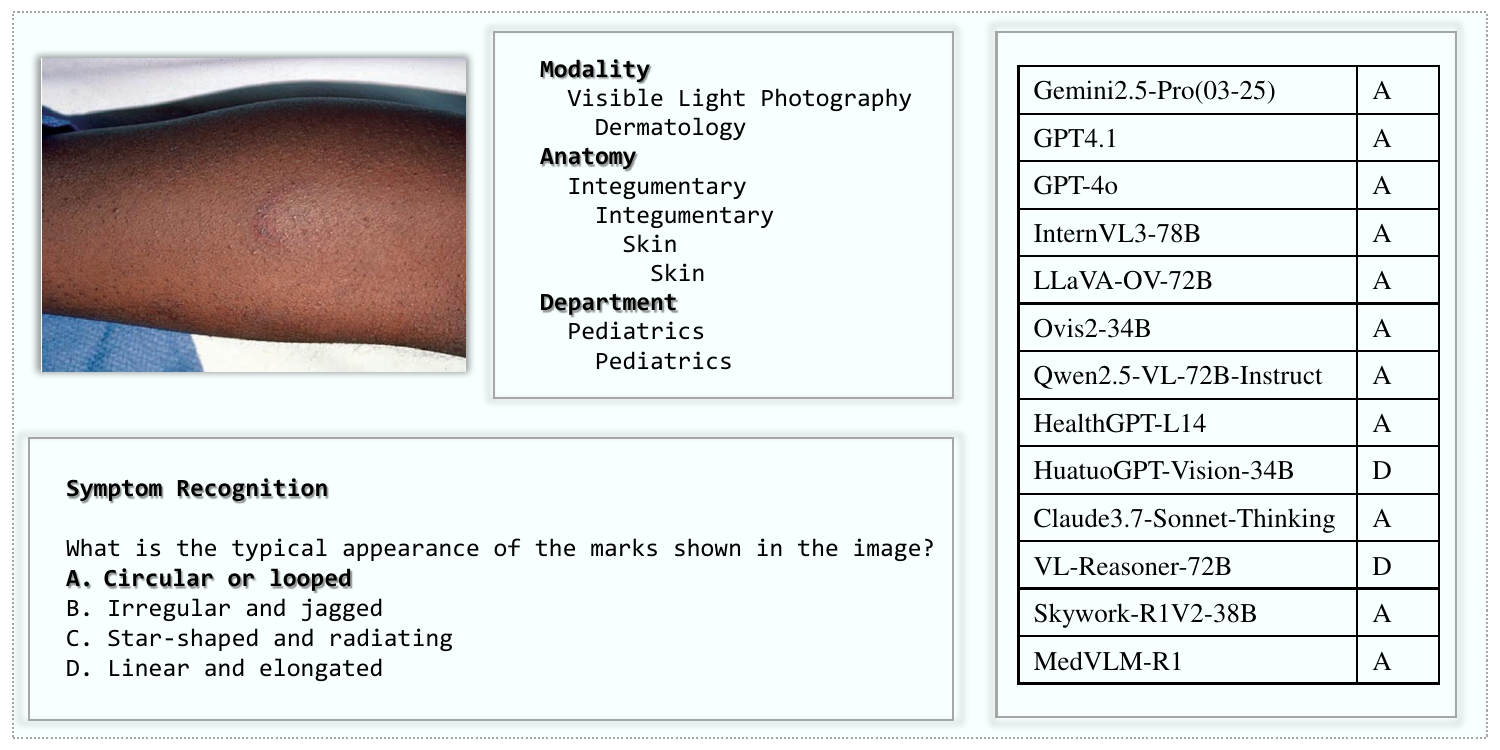}
\label{Entry_BodyPart_Skin}
\end{figure}

\begin{figure}[!th] 
\centering
\includegraphics[width=1\textwidth]{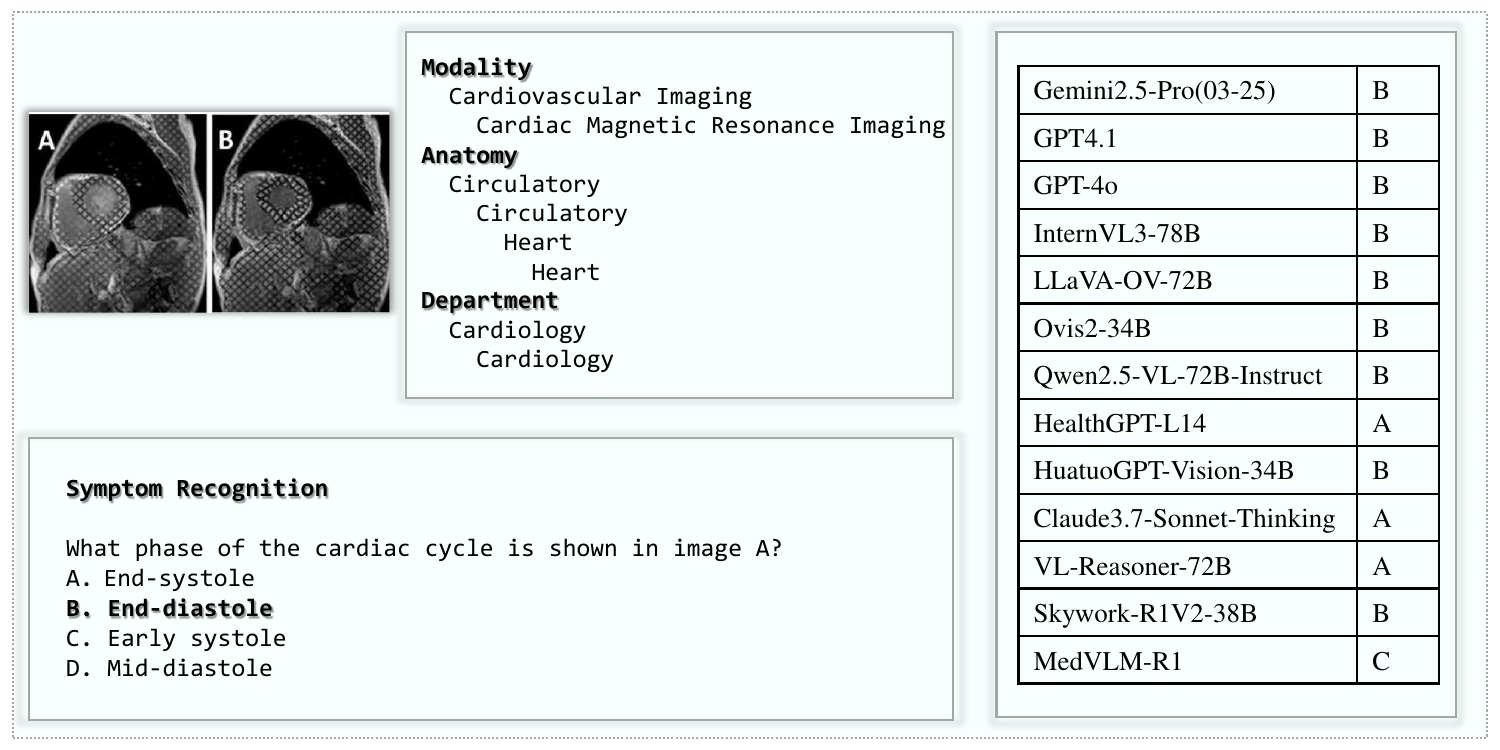}
\label{Entry_Department_Cardiology}
\end{figure}

\newpage
\begin{figure}[!th] 
\centering
\includegraphics[width=1\textwidth]{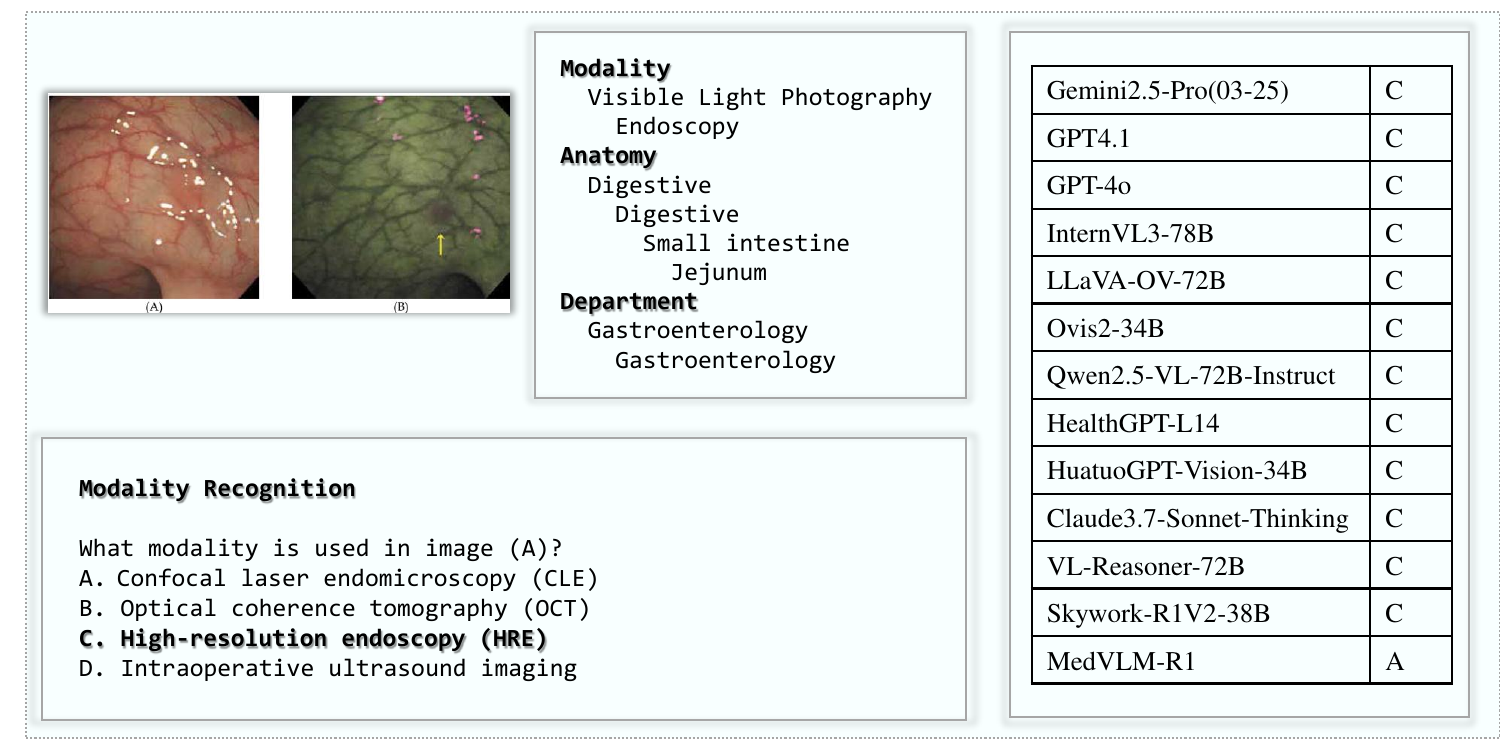}
\label{Entry_Department_Gastroenterology}
\end{figure}

\begin{figure}[!th] 
\centering
\includegraphics[width=1\textwidth]{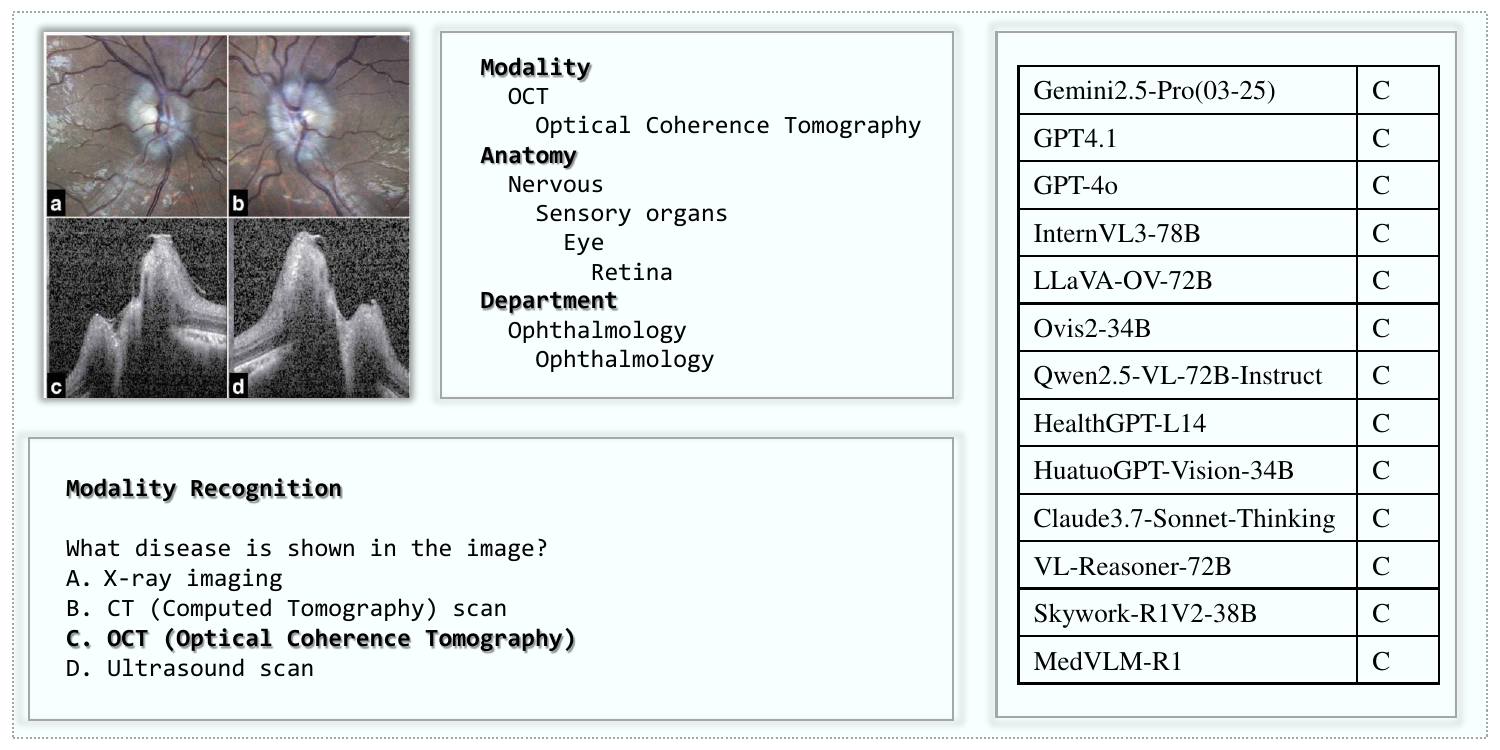}
\label{Entry_BodyPart_Eye}
\end{figure}

\newpage
\begin{figure}[!th] 
\centering
\includegraphics[width=1\textwidth]{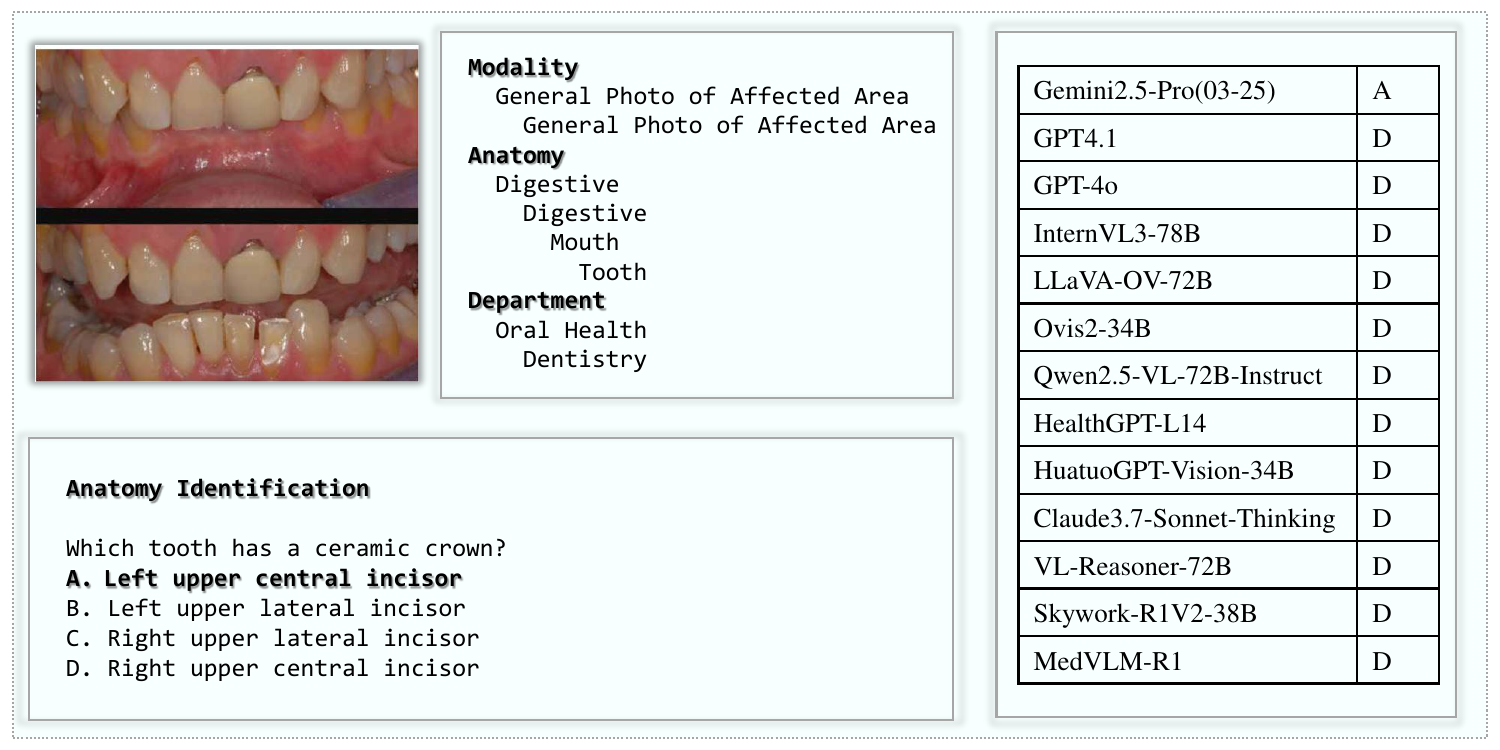}
\label{Entry_BodyPart_Mouth}
\end{figure}

\begin{figure}[!th] 
\centering
\includegraphics[width=1\textwidth]{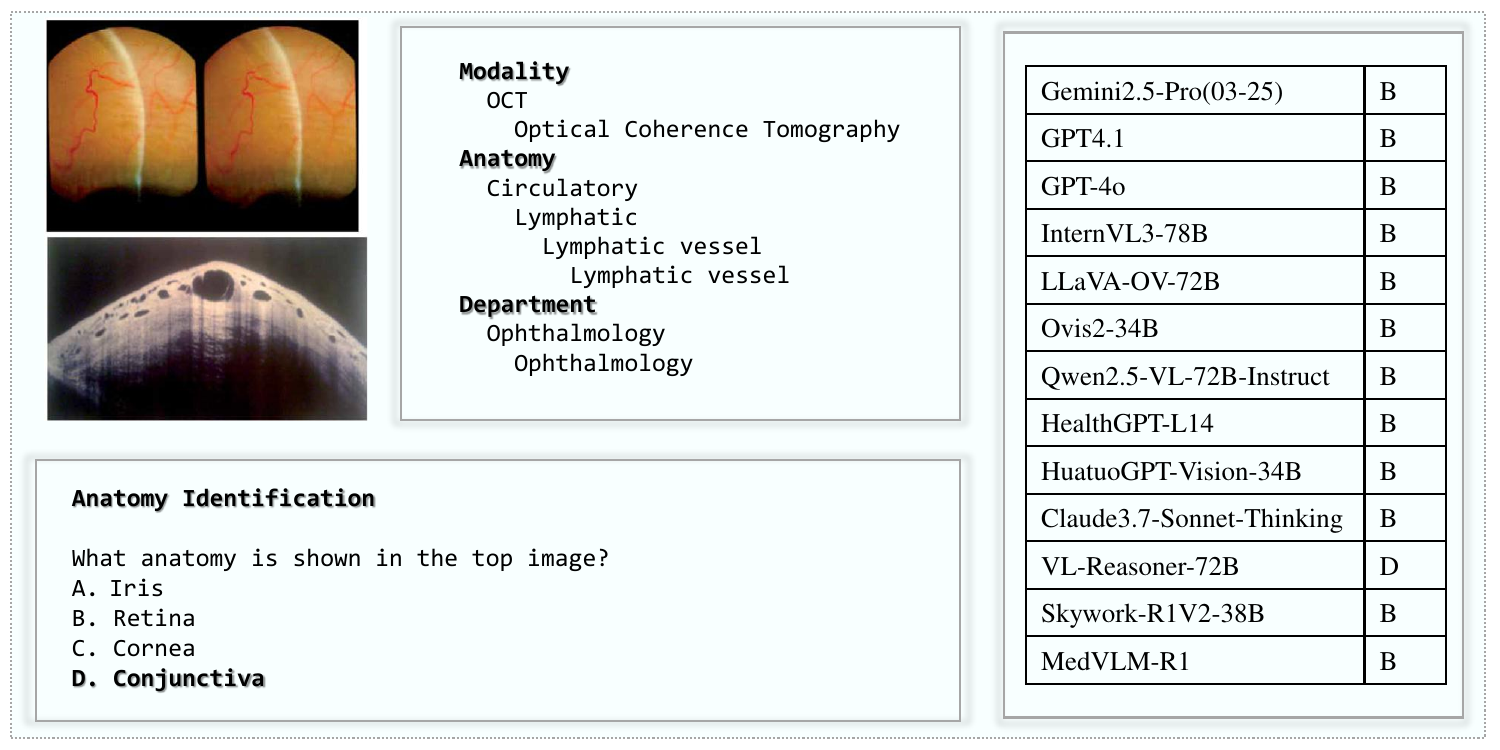}
\label{Entry_Department_Ophthalmology}
\end{figure}

\newpage
\begin{figure}[!th] 
\centering
\includegraphics[width=1\textwidth]{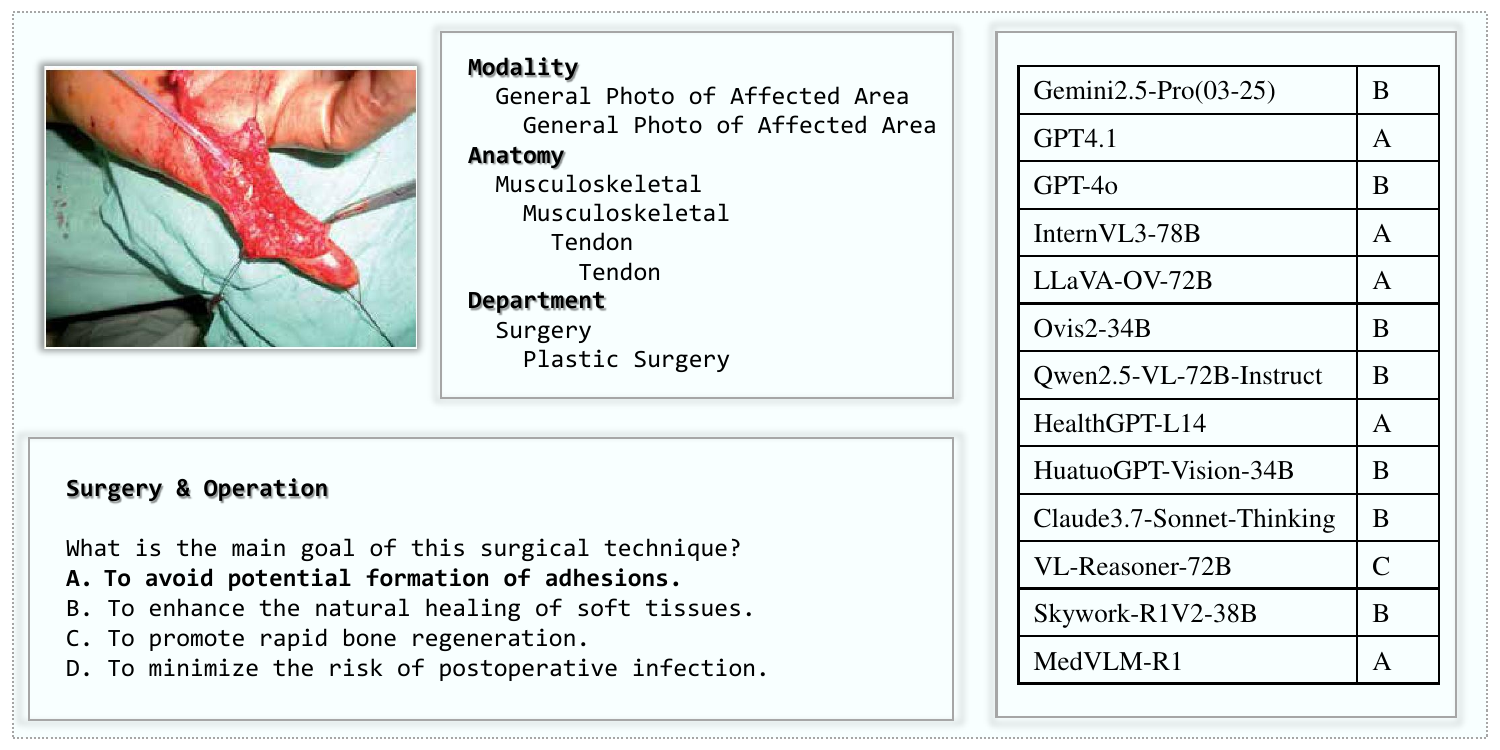}
\label{Entry_Department_PlasticSurgery}
\end{figure}

\begin{figure}[!th] 
\centering
\includegraphics[width=1\textwidth]{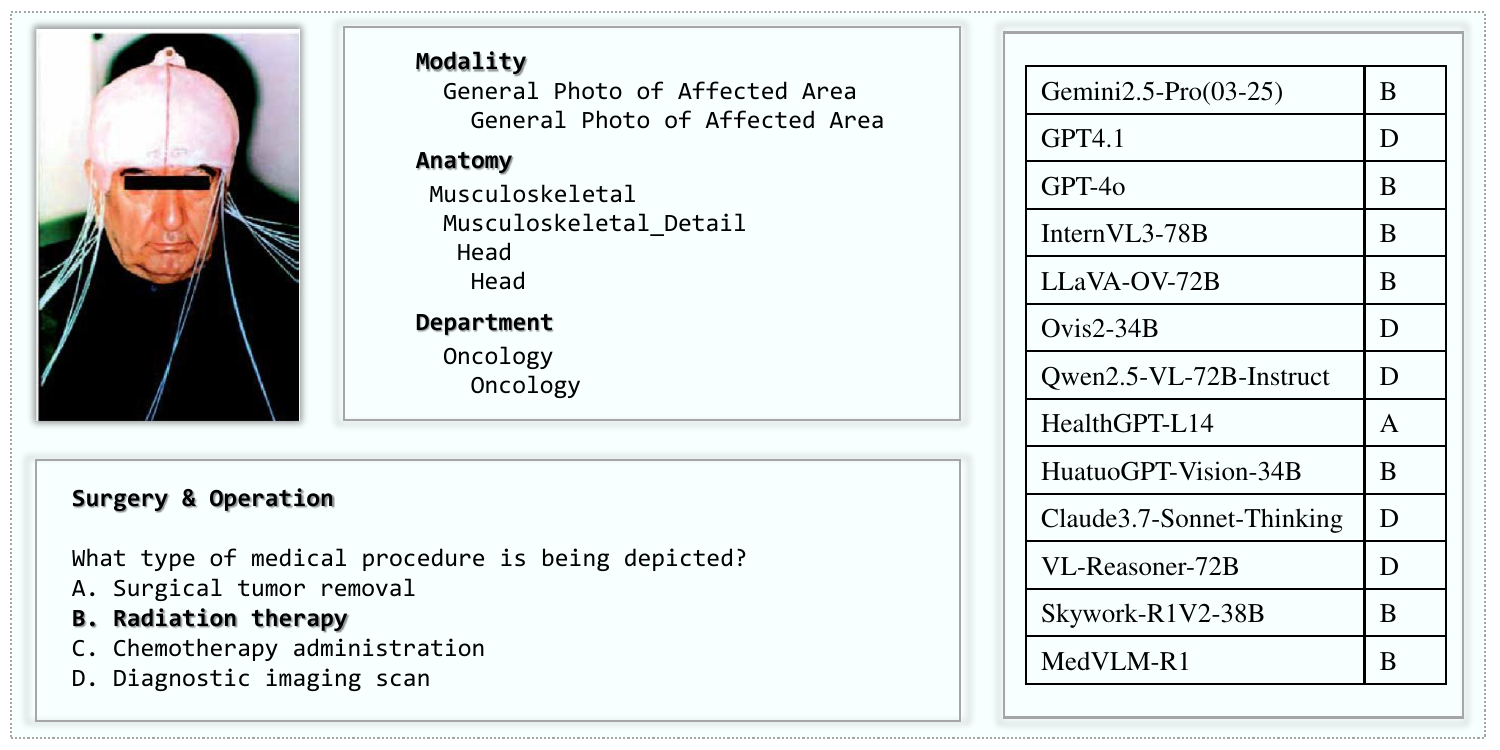}
\label{Entry_BodyPart_Head}
\end{figure}

\end{document}